\documentclass[preprint,3p,times,twocolumn]{elsarticle}

\usepackage{multirow}
\usepackage{hhline}
\usepackage{url}
\usepackage{amssymb}
\usepackage{amsmath}
\usepackage[noend]{algpseudocode}
\usepackage{algorithmicx,algorithm}
\usepackage { hyperref } 
\hypersetup { colorlinks,allcolors=black }
\usepackage{amssymb}

\journal{arXiv}

\begin{document}

\begin{frontmatter}



\title{MC-LCR: Multi-modal contrastive classification by locally correlated representations for effective face forgery detection}


\address[1]{Engineering Research Center of Cyberspace, Yunnan University, Kunming, PR China}
\address[2]{School of Software, Yunnan University, Kunming, PR China}
\address[3]{School of Artificial Intelligence and Computer Science, Jiangnan University, Wuxi, PR China}
\address[4]{School of Cyber Science and Engineering, Wuhan University, Wuhan, PR China}

\author[1,2]{Gaojian Wang}
\author[1,2]{Qian Jiang}
\author[1,2]{Xin Jin}
\author[3]{Wei Li}
\author[4,]{Xiaohui Cui}\fntext[label2]{Corresponding author. 
E-mail address: xcui@whu.edu.cn}

\begin{abstract}
As the remarkable development of facial manipulation technologies is accompanied by severe security concerns, face forgery detection has become a recent research hotspot. Most existing detection methods train a binary classifier under global supervision to judge real or fake. However, advanced manipulations only perform small-scale tampering, posing challenges to comprehensively capture subtle and local forgery artifacts, especially in high compression settings and cross-dataset scenarios. To address such limitations, we propose a novel framework named \textit{Multi-modal Contrastive Classification by Locally Correlated Representations} (MC-LCR), for effective face forgery detection. Instead of specific appearance features, our MC-LCR aims to amplify implicit local discrepancies between authentic and forged faces from both spatial and frequency domains. Specifically, we design the \textit{shallow style representation block} that measures the pairwise correlation of shallow feature maps, which encodes local style information to extract more discriminative features in the spatial domain. Moreover, we make a key observation that subtle forgery artifacts can be further exposed in the patch-wise phase and amplitude spectrum and exhibit different clues. According to the complementarity of amplitude and phase information, we develop a \textit{patch-wise amplitude and phase dual attention} module to capture locally correlated inconsistencies with each other in the frequency domain. Besides the above two modules, we further introduce the collaboration of supervised contrastive loss with cross-entropy loss. It helps the network learn more discriminative and generalized representations. Through extensive experiments and comprehensive studies, we achieve state-of-the-art performance and demonstrate the robustness and generalization of our method.
\end{abstract}




\end{frontmatter}


\section{Introduction}
Visual forgeries are now rampant on social media and the Internet, with the remarkable progress of deep generative models, such as Variational Autoencoders\cite{pu2016variational, kingma2013auto} and Generative Adversarial Networks\cite{goodfellow2014generative}. Nowadays, even non-specialists can easily manipulate facial content through booming apps\cite{faceapp,deepfakesweb} and open-source tools\cite{DeepFakes, FaceSwap, DeepFaceLab}. Although these neutral synthesis techniques can generate high-quality images or videos to help create facial visual effects\cite{avatarify}, they can also be abused by malicious users. Misinformation tends to spread quickly on the Internet, causing severe trust and security concerns in our society\cite{kwok2021deepfake, pantserev2020malicious}, given that the personal face implies sensitive identity information. To against various facial manipulations in practical scenarios, developing more robust and generalized face forgery detection methods is a paramount and urgent need.

Benefiting from the breakthrough of deeper and deeper neural networks in image classification\cite{chollet2017xception, tan2019efficientnet}, DNN-based detection methods have sprung up\cite{afchar2018mesonet, nguyen2019capsule, rossler2019faceforensics++, guera2018deepfake, sabir2019recurrent}. Most of these models follow the backbone network stream to extract the entire facial image's global features directly and then make the decision through a binary classifier. However, some advanced manipulation methods\cite{thies2019deferred,li2020celeb} only produce subtle forgery traces locally, resulting in the discriminability of global features suffering from small-scale tampering. Recent works observe this phenomenon and suggest exploring discrepancies in local regions, such as creating an attention map to indicate manipulated regions\cite{dang2020detection}, focusing on local patches with patch-based classifiers\cite{chai2020makes}, and attending to different local parts by multiple spatial attention heads\cite{zhao2021multi}. Although some of the methods mentioned above emphasize local features in the spatial domain, the forgery artifacts in color-space are fragile to compression, thus limiting their robustness in high compression settings.

Meanwhile, frequency domain information introduces another perspective for mining forged clues. Prior studies\cite{durall2020watch, frank2020leveraging} have found that the up-sampling structure of generators leads to checkerboard artifacts in the frequency domain. They usually use Fourier transform\cite{durall2019unmasking, zhang2019detecting} or Discrete Cosine Transform\cite{frank2020leveraging} to acquire frequency spectrums and then feed them to the detector. Liu et al.\cite{liu2021spatial} further prove that the phase spectrum is more sensitive to up-sampling than the amplitude spectrum. However, their performance drops greatly when encountering manipulation methods without the up-sampling step. What is more, most of the existing frequency-based methods directly transform the entire image into the frequency spectrum for analysis. Some local tampered faces\cite{thies2016face2face, thies2019deferred} show indiscernible anomalies in the global frequency domain, such that these methods also suffer from subtle local artifacts. F3-Net\cite{qian2020thinking}, which is consisted of two frequency-aware branches, keeps the importance of local frequency information in mind and achieves state-of-the-art performance in highly compressed videos. However, it cannot guarantee generalization only by mining subtle frequency patterns.

Based on these observations of current face forgery and detection methods, the key motivations behind our works are mainly: a) In addition to the high-level semantic features, subtle tampering traces should also be effectively captured in the spatial domain. Moreover, forged faces may expose similar face warping\cite{li2018exposing} and swapping artifacts\cite{li2020face} across different areas. Correlated encoding of these local patterns enables concentration more on commonly counterfeit evidence than a specific manipulation method. b) The limited pixels modified by small-scale tampering methods also do not introduce distinct frequency component abnormalities, thus forged clues mined in the frequency domain should also tend to be localized. On the other hand, it is incomprehensive to utilize only the amplitude\cite{durall2019unmasking} or phase\cite{liu2021spatial} information since they complement each other, and both should be jointly exploited. c) Most DNN-based models treat face forgery detection as a vanilla binary classification problem and learn end-to-end with the supervision of cross-entropy loss. However, training a generalizable detection model with so many multi-modal features deeply mined from both spatial and frequency domains is a nontrivial task. This is mainly because some subtle artifacts are manipulation-specific, and binary cross-entropy loss does not directly encourage wider separation between cross-domain forged faces, i.e., overfitting. 

Inspired by the above thoughts, we propose a novel Multi-modal Contrastive Classification by Locally Correlated Representations (MC-LCR) for face forgery detection. Corresponding to the aforementioned motivations, our framework is composed of three meticulously devised modules. The first is the \textit{shallow style representation block} (SSRB), which aims at extracting low-level style features from the spatial domain. To hold subtle artifacts in local regions, SSRB truncates the feature maps from shallow layers and measures the pairwise correlation of feature maps by refining the Gram matrix\cite{gatys2015texture}. Secondly, we make a crucial observation that although forged discrepancies are minimal in color-space and entire frequency spectrums, they still can be revealed in spectrums of local patches. Specifically, patch-wise amplitude spectrums expose more visible checkboard artifacts, and patch-wise phase spectrums exhibit more differences. Since amplitude and phase are complementary frequency information, we further propose \textit{Patch-wise Amplitude and Phase Dual Attention module} (PAPDA) to guide interactions with each other, which aims to capture locally correlated inconsistencies in amplitude and phase spectrum. Then the frequency branch network composes more abundant and informative amplitude and phase features. Thirdly, these multi-modal features are fused and then projected into an additional embedding space. In such a restricted space, the supervised contrastive loss\cite{khosla2020supervised} explicitly encourages facial embeddings of the same class to be put together while increasing the distance between different categories. With the collaboration of the classification head supervised by cross-entropy loss, our MC-LCR network learns a discriminative and generalized locally correlated representation from multi-modality.

To demonstrate the effectiveness of the proposed MC-LCR, we conduct comprehensive evaluations on the FF++\cite{rossler2019faceforensics++} dataset with different compression settings and manipulation methods. Moreover, we achieve state-of-the-art performance on the FF++ while still keeping superior generalization on the cross-dataset evaluation of challenging Celeb-DF\cite{li2020celeb} and DeeperForensics\cite{jiang2020deeperforensics}. In summary, the contributions of this work are presented as follows:

$\bullet$ We propose a novel framework for effective and robust face forgery detection, named Multi-modal contrastive classification by locally correlated representations, which mines local discriminative features from different modalities. 

$\bullet$ SSRB and PAPDA are developed to capture subtle artifacts in the spatial and frequency domain, respectively. We also introduce the combination of supervised contrastive loss and cross-entropy loss to learn generalizable representations.

$\bullet$ We perform a frequency analysis on local image patches and reveal the implicit frequency discrepancy on the patch-wise amplitude and phase spectrum. We collaborate amplitude and phase information based on the proposed PAPDA to model their correlation interactively.

$\bullet$ Extensive experiments demonstrate the effectiveness of our method for face forgery detection. Ablation studies and visualizations are presented to give new insights into our state-of-the-art performance.

\section{Related work}
Face forgery detection is a classic problem in computer vision and image forensics\cite{farid2009image, piva2013overview}. Recently, with the remarkable progress of computer graphics and deep generative models, conventional image forgery detection methods struggle to distinguish realistic deepfakes. As face forgery has gained more and more attention, various countermeasures have been proposed to tackle the increasing challenges. Nevertheless, improved synthesis algorithms, small-scale tampering, and high compression settings still pose threats to current detection methods in terms of effectiveness, generalization, and robustness. In this section, we briefly review these previous works.

\subsection{Spatial-based forgery detection methods}
Earlier facial manipulation techniques are prone to produce obviously forged artifacts\cite{korshunov2018deepfakes, li2018ictu}, which inspired many detection methods to explore anomalies in the spatial domain, such as lack of eye blinking\cite{li2018ictu}, abnormal head pose\cite{yang2019exposing}, face warping artifacts\cite{li2018exposing}, and visual artifacts\cite{matern2019exploiting}. However, the improved generation algorithms\cite{li2020celeb} significantly reduce visible artifacts, rendering these methods powerless. Face X-ray\cite{li2020face} further observed the blending step of the face swap and located the resulting boundaries, which significantly improved the generalization performance. However, its performance suffers from the weakened boundaries in highly compressed video scenarios, and it may not be extrapolated to entire face synthesis without a blending step.

Turning to the excellent automatic feature extraction capabilities of deep neural networks, many DNN-based methods take RGB pixels as inputs and treat face forgery detection as a Vanilla binary classification problem. MesoNet\cite{afchar2018mesonet} introduced two CNN architectures to detect forged faces based on the mesoscopic properties of images. Nguyen et al.\cite{nguyen2019capsule} combined the capsule network with the pre-trained VGG19 model for face forensics. Rossler et al.\cite{rossler2019faceforensics++} released a benchmark for facial manipulation detection and showed the effectiveness of Xception\cite{chollet2017xception}. Considering the temporal information in the video stream, some works adopted recurrent neural network models, such as LSTM\cite{guera2018deepfake} and GRU\cite{sabir2019recurrent}. 

Instead of blindly utilizing deep learning directly as the detector, a series of methods focus on forgery artifacts. Dang et al.\cite{dang2020detection} proposed to highlight the forged regions and locate manipulations through learned attention maps. Liu et al.\cite{liu2020global} leveraged the global texture information to detect the images entirely generated by GANs, but it seemed weak in the face of the locally tampered deepfakes. Recently, Zhao et al.\cite{zhao2021multi} presented the MADD framework for deepfake detection, exploring different local parts guided by multiple attention maps and combining enhanced shallow texture features. However, most of these spatial-based detection methods are susceptible to compression settings, given that subtle manipulation traces are quality-sensitive. Besides, forgery artifacts take various forms in the color-space, which also challenges the generalization ability of these models. On the other hand, many works on mining frequency clues of forged faces have been proposed.

\begin{table*}[!htb]
\centering
\caption{Limitation of existing forgery detection methods and the novelty of the proposed MC-LCR.}
\label{tab:com}
\resizebox{\linewidth}{!}{%
\begin{tabular}{c|c|c|l} 
\hline
\textbf{Categories}                                                        & \textbf{Methods}                                        & \multicolumn{2}{c}{\textbf{Limitations}}       \\ 
\hline
\multirow{11}{*}{\begin{tabular}[c]{@{}c@{}}spatial-\\based\end{tabular}}  & eye blinking\cite{li2018ictu} & \textbf{a-c-d-e-l-} & \multirow{18}{*}{\begin{tabular}[c]{@{}l@{}}\textbf{a-} Limit to specific forgery artifacts or hand-craft features and lack effectiveness for advanced deepfakes with\\ higher quality.\\\textbf{ b-} The detector is trained only via the global spatial features extracted by CNN and suffers from small-scale\\ manipulations that only produce subtle forgery traces locally.\\\textbf{ c-} The direct use of naïve DNN models for detection is susceptible to overfitting and lacks generalization ability.\\\textbf{ d-} The spatial forgery artifacts in color-space are fragile to compression, thus limiting their robustness in high\\ compression settings.\\\textbf{ e-} The extracted features take various forms in the color-space, which challenges the generalization ability.\\\textbf{ f-} It identifies checkerboard artifacts or frequency component anomalies in the frequency domain, but the\\ performance drops significantly against local forgeries or manipulations without the upsampling step.\\\textbf{ g-} Frequency-based methods that transform the entire image into the frequency spectrum for analysis and\\ detection, but some local tampered faces show indiscernible anomalies in the global frequency domain.\\\textbf{ h-} It cannot guarantee generalization only by mining subtle frequency patterns.\\\textbf{ i-} The use of fixed filters or hand-crafted features to explore frequency information limits discernibility.\\\textbf{ j-} It is incomprehensive to utilize only the amplitude or phase information since they complement each other.\\\textbf{ k-} Information in the spatial domain is not utilized.\\\textbf{ l-} The model only performs classification learning under supervision such as cross-entropy loss and does not\\ directly encourage wider separation between cross-domain forged faces, i.e., overfitting and lack generalizability.\\\end{tabular}}  \\
& head pose\cite{yang2019exposing}& \textbf{a-c-d-e-l-} &\\ 
& FWA\cite{li2018exposing}& \textbf{a-c-d-e-l-} & \\
& VA\cite{matern2019exploiting}& \textbf{a-c-d-e-l-} & \\ 
& Face X-ray\cite{li2020face}& \textbf{d-l-}&    \\
& Mesonet\cite{afchar2018mesonet}& \textbf{b-c-d-e-l-} & \\
& Capsule\cite{nguyen2019capsule}& \textbf{b-c-d-e-l-} &\\
& Xception\cite{rossler2019faceforensics++}& \textbf{b-c-d-e-l-} &  \\
& Dang et al.\cite{dang2020detection}& \textbf{d-}         &\\
& Liu et al.\cite{liu2020global}& \textbf{b-l-}&\\
& MADD\cite{zhao2021multi}& \textbf{d-e-}       &\\
\cline{1-3}
\multirow{4}{*}{\begin{tabular}[c]{@{}c@{}}frequency-\\based\end{tabular}} & Durall et al.\cite{durall2019unmasking}                                           & \textbf{f-g-j-k-l-} &\\
& AutoGAN\cite{zhang2019detecting}& \textbf{f-g-j-k-l-}   &\\
                                                                           & Frank et al.\cite{frank2020leveraging}                                            & \textbf{f-g-k-l-}     &\\
                                                                           & F3-Net\cite{qian2020thinking}                                                  & \textbf{h-k-l-}     &\\ 
\cline{1-3}
\multirow{4}{*}{Combined}                                                  & SSTNet\cite{wu2020sstnet}                                                  & \textbf{i-l-}       &\\
                                                                           & Two-branch\cite{masi2020two}                                              & \textbf{i-l-}       &\\
                                                                           & SPSL\cite{liu2021spatial}                                                    & \textbf{j-l-}       &\\ 
\cline{2-4}
                                                                           & \begin{tabular}[c]{@{}c@{}}MC-LCR\\ (ours)\end{tabular} & \textbf{Novelty→}   & \begin{tabular}[c]{@{}l@{}}To \textbf{c-d-e-k}:\\ MC-LCR concentrates on the local discrepancies between authentic and forged faces from both spatial and\\ frequency domains. It consists of SSRB, PAPDA, and contrastive classification components, which aim to\\ improve the effectiveness, robustness, and generalization of face forgery detection.\\ To \textbf{b-h-}:\\ Inspired by the low-level texture artifacts introduced by different manipulation methods are more discriminative\\ and generalizable in smaller receptive fields, the proposed SSRB encodes local style representation from shallow\\ layers.\\ To \textbf{f-g-j-}:\\ The proposed PAPDA interactively captures implicated checkerboard artifacts and differences in the patch-wise\\ amplitude and phase spectrum, to extract robust local frequency features for detection.\\ To \textbf{a-i-}:\\ MC-LCR is an end-to-end framework where all components are fully learning-driven.\\ To \textbf{l-}:\\ The collaboration of supervised contrastive loss with cross-entropy loss learns more separate encoding\\ representations for multi-modal features, further improving generalization ability.\\\end{tabular}\\
\hline
\end{tabular}}
\end{table*}

\subsection{Frequency-based forgery detection methods}
In the field of digital signal processing and computer vision, frequency domain analysis has always been a classical and powerful method, which has also been exploited for forgery image detection. In fact, the up-sampling step of the deep generative models will introduce anomalies in the frequency domain, such as distribution differences of high-frequency components\cite{durall2020watch} and checkerboard artifacts\cite{frank2020leveraging}. 

Durall et al.\cite{durall2019unmasking} averaged the amplitude spectrum with DFT and used distribution differences to distinguish fake images. AutoGAN\cite{zhang2019detecting} simulated the checkerboard artifacts produced by GAN in the frequency domain and input the DFT spectrum into the detector. Frank et al.\cite{frank2020leveraging} transformed the image from spatial to frequency domain through DCT and analyzed the frequency anomalies of various GANs. These methods are indeed effective for detecting images synthesized entirely by GAN. However, some local manipulations\cite{thies2019deferred} will not cause apparent abnormalities in the global frequency domain, and the performance of these detection methods on small-scale tampering drops greatly. F3-Net\cite{qian2020thinking} reported state-of-the-art detection results on high compression deepfake videos, which designed frequency-aware decomposition and local frequency statistics to mine subtle forgery patterns in the frequency domain. Nevertheless, F3-NET showed unsatisfactory generalization ability on cross-dataset evaluation.

\subsection{Forgery detection methods combining spatial and frequency features}
To extract features comprehensively and complementary, considering both spatial and frequency domain information gradually becomes the research mainstream\cite{wu2020sstnet, masi2020two, li2021frequency, luo2021generalizing, liu2021spatial, chen2021local}.
SSTNet\cite{wu2020sstnet} detected manipulated face images by extracting spatial, steganalysis, and temporal features with modified Xception and LSTM. Two-branch RN\cite{masi2020two} combined information from both spatial and frequency domains, and a Laplacian of Gaussian (LOG) was used to enhance multi-band frequencies. However, the use of fixed filters or hand-crafted features limits discernibility. Recently, SPSL\cite{liu2021spatial} has observed that the up-sampling structure will cause more significant pixel differences in the phase spectrum than amplitude spectrum, and combined phase information with shallow feature maps to improve the generalizability of face forgery detection. Nevertheless, SPSL also struggles to capture small-scale frequency artifacts caused by local manipulation methods in the global phase spectrum, limiting its in-domain evaluation performance and robustness.

This section reviews the state-of-the-art face forgery detection methods, and a summary of the limitation of existing forgery detection methods and the novelty of the proposed method is provided in Table \ref{tab:com}. The detailed comparison clarifies the contributions of this work and provides insights for future research and exploration.

\begin{figure*}[htb]
  \centering
  \includegraphics[width=1\linewidth]{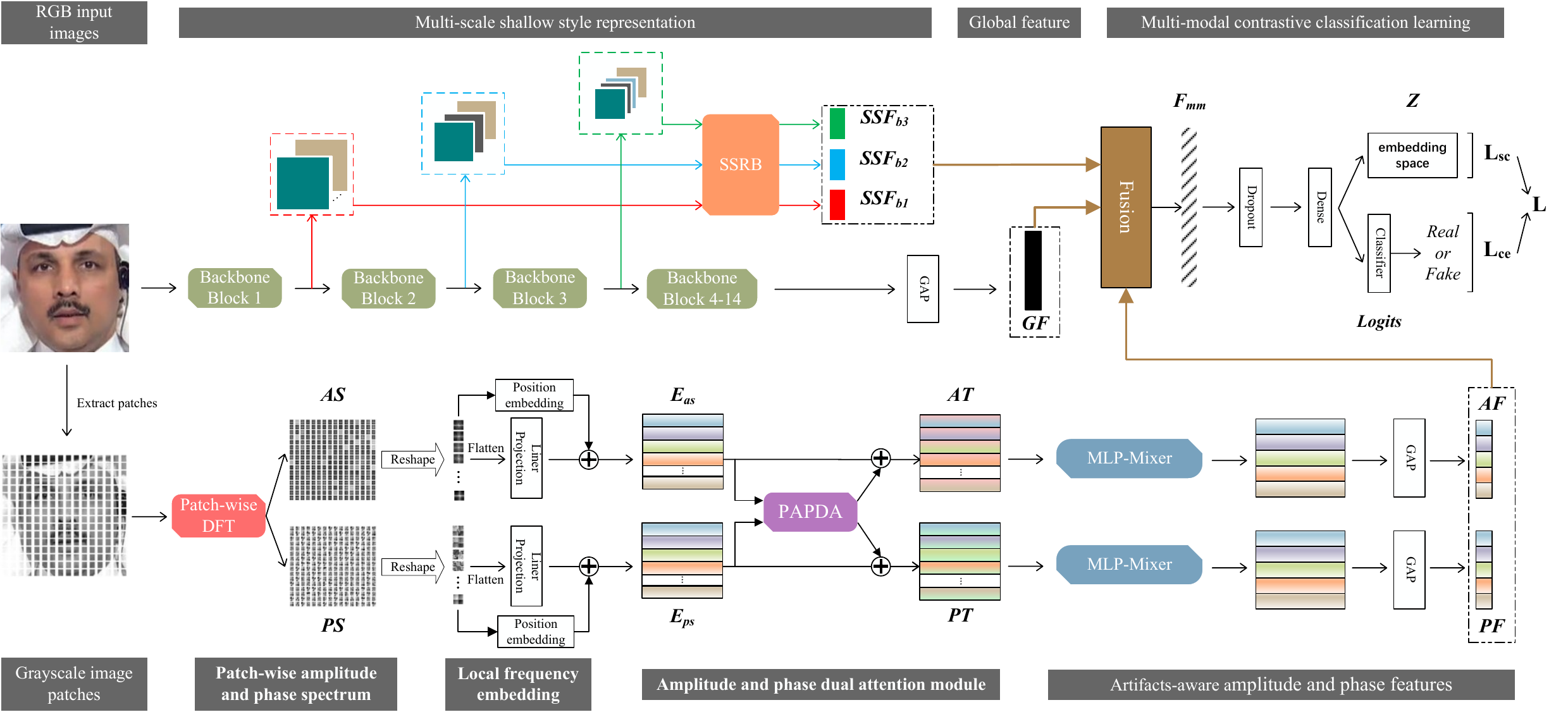}
  \caption{The framework of our MC-LCR. We present a two-stream architecture to work in the spatial and frequency domains. The spatial branch extracts shallow style features($\textit{SSF}_{b1}$, $\textit{SSF}_{b2}$, and $\textit{SSF}_{b3}$) and global features($\textit{GF}$) from SSRB and backbone network, respectively. The frequency branch processes amplitude features($\textit{AF}$) and phase features($\textit{PF}$). These multi-modal features are fused as a rich representation $F_{MM}$ for contrastive learning and classification. We train the framework in an end-to-end manner with the joint of supervised contrastive loss($L_{SC}$) and cross-entropy loss($L_{CE}$). SSRB represents the shallow style representation block that truncates the feature maps from shallow layers. PAPDA stands for the patch-wise amplitude and phase dual attention module. GAP represents the global average pooling layer.}
\label{fig:Framework}
\end{figure*}

\section{Methods} \label{sec:Methods}

In this section, we propose the Multi-modal Contrastive Classification by Locally Correlated Representations (MC-LCR) for face forgery detection, as illustrated in Figure \ref{fig:Framework}. To tease the framework systematically, we first briefly introduce the main structure and algorithmic workflow of MC-LCR in Section \ref{sec:ov}, and then present the design motivation, the structure of proposed modules, and details of the MC-LCR: We present the proposed SSRB in Section \ref{sec:SSRB}; We analyze the amplitude and phase anomalies of the forged faces in Section \ref{sec:PWAPS} and further design the PAPDA in Section \ref{sec:PAPDA}; Finally, we introduce the contrastive classification learning via multi-modal features in Section \ref{sec:MFCC}.

\subsection{Overview}\label{sec:ov}

\begin{algorithm*}[h!]
\renewcommand{\thealgorithm}{1}
\caption{Main algorithm of training MC-LCR}
{\bf Input:}
Training dataset $T$=$\{x, y\}$ with image data $x$ and label $y$;\\
Validation dataset $V$;\\
 Hyperparameters: patch size $P$, mini batch size, training epoch $E$, loss weight $\alpha$. \\
{\bf Output:} 
Optimal MC-LCR model for face forgery detection.
\begin{algorithmic}[1]
\State Initialize the parameters of $\textit{Xception}$ with imagenet pretrained weights;
\State Initialize the parameters of other modules (SSRB, PAPDA, MLP-Mixer) and layers ($\textit{Dense}$, $\textit{Embedding}$) on MC-LCR model;
\For{$epoch$ \textbf{in} $E$}
\For{Image $I$ \textbf{in} $mini\_batch \in T$}
\State $\it SSF_{b1}$ $\leftarrow$ SSRB(\textit{Xception\_block1}($I$));
\State $\it SSF_{b2}$ $\leftarrow$ SSRB(\textit{Xception\_block2}($I$));
\State $\it SSF_{b3}$ $\leftarrow$ SSRB(\textit{Xception\_block3}($I$));
\State $GF$ $\leftarrow$ Xception($I$);
\State Convert image $I$ to gray-scale, and then split it to $N$ patches $\left\{I_{i}\right\}_{i=1}^{N}$ with size $p\text{x}p$;
\For{$i$ \textbf{in} $N$}
\State $AS_{i}$, $PS_{i}$ $\leftarrow$ patch-wise $DFT$($I_{i}$);
\EndFor
\State $E_{as}$ $\leftarrow$ $\it Dense$($\left\{AS_{i}\right\}_{i=1}^{N}$) + $\it Position\_Embedding$($\left\{AS_{i}\right\}_{i=1}^{N}$);
\State $E_{ps}$ $\leftarrow$ $\it Dense$($\left\{PS_{i}\right\}_{i=1}^{N}$) + $\it Position\_Embedding$($\left\{PS_{i}\right\}_{i=1}^{N}$);
\State $AT$, $PT$ $\leftarrow$ PAPDA($E_{as}$, $E_{ps}$);
\State $AF$ $\leftarrow$ $\it GAP$(MLP-Mixer($AT$));
\State $PF$ $\leftarrow$ $\it GAP$(MLP-Mixer($PT$)); 
\State $F_{mm}$ = $concatenate$($\it SSF_{b1}$, $\it SSF_{b2}$,$\it SSF_{b3}$, $GF$, $AF$, $PF$);
\State $F_{E}$ $\leftarrow$ $\it Dense$($F_{mm}$);
\State $z$ = $\it Dense$($F_{E}$);
\State $preds$ = $\it Dense\_classifier$($F_{E}$);
\State $loss$ = $\alpha L_{sc}$($z$, $y$) + $\alpha L_{ce}$($preds$, $y$);
\EndFor
\State Update all parameters of MC-LCR to minimize $loss$ by $Adam$;
\EndFor
\State \textbf{Return:} The MC-LCR model with $weights$ $\leftarrow$ argmax($ACC\_score$(MC-LCR($V$))).
\end{algorithmic}
\label{MC-LCR}
\end{algorithm*}

As shown in Figure \ref{fig:Framework}, the proposed MC-LCR is a two-stream framework, which works in both spatial and frequency domains. In the spatial branch, Xception\cite{chollet2017xception} is employed as backbone network. The proposed SSRB (\textit{shallow style representation block}) truncates feature maps from blocks 1-3 of Xception, and then extracts shallow style features $\it SSF_{b1}$, $\it SSF_{b2}$, and $\it SSF_{b3}$, respectively. These local features served to complement the global semantic feature $GF$ output by the last block of Xception. In the frequency branch, the input image $I$ is first converted to grayscale and segmented into patches $\left\{I_{i}\right\}_{i=1}^{N}$. Then, each patch $I_{i}$ is separately performed Discrete Fourier Transform, i.e., patch-wise DFT, to obtain the amplitude spectrum $AS_{i}$ and phase spectrum $PS_{i}$. The patch-wise frequency spectrums $\left\{AS_{i}\right\}_{i=1}^{N}$ and $\left\{PS_{i}\right\}_{i=1}^{N}$ are flattened and then projected linearly to frequency embeddings $E_{as}$ and $E_{ps}$, respectively, which have been encoded position relationship by an additional embedding layer. After that, the proposed PAPDA (\textit{Patch-wise amplitude and phase dual attention module}) interactively captures frequency anomaly patterns from $E_{as}$ and $E_{ps}$, and generates amplitude tokens $AT$ and phase tokens $PT$. Finally, the refined local amplitude features $AF$ and phase features $PF$ are extracted by MLP-Mixer network and the last global average pooling (GAP) layer. 

These multi-modal representations ($\it SSF_{b1}$, $\it SSF_{b2}$, $\it SSF_{b3}$, $GF$, $AF$, $PF$) extracted from different branches are concatenated to produce $F_{mm}$. After $F_{mm}$ is encoded as $F_{E}$, contrastive learning and classification learning are performed in the embedding space and the classifier, where the training procedure is guided by supervised contrastive loss $L_{sc}$ and cross-entropy loss $L_{ce}$, respectively. The overall learning algorithm of MC-LCR is presented in Algorithm \ref{MC-LCR}.

\begin{figure*}[!htb]
  \centering
  \includegraphics[width=.9\linewidth]{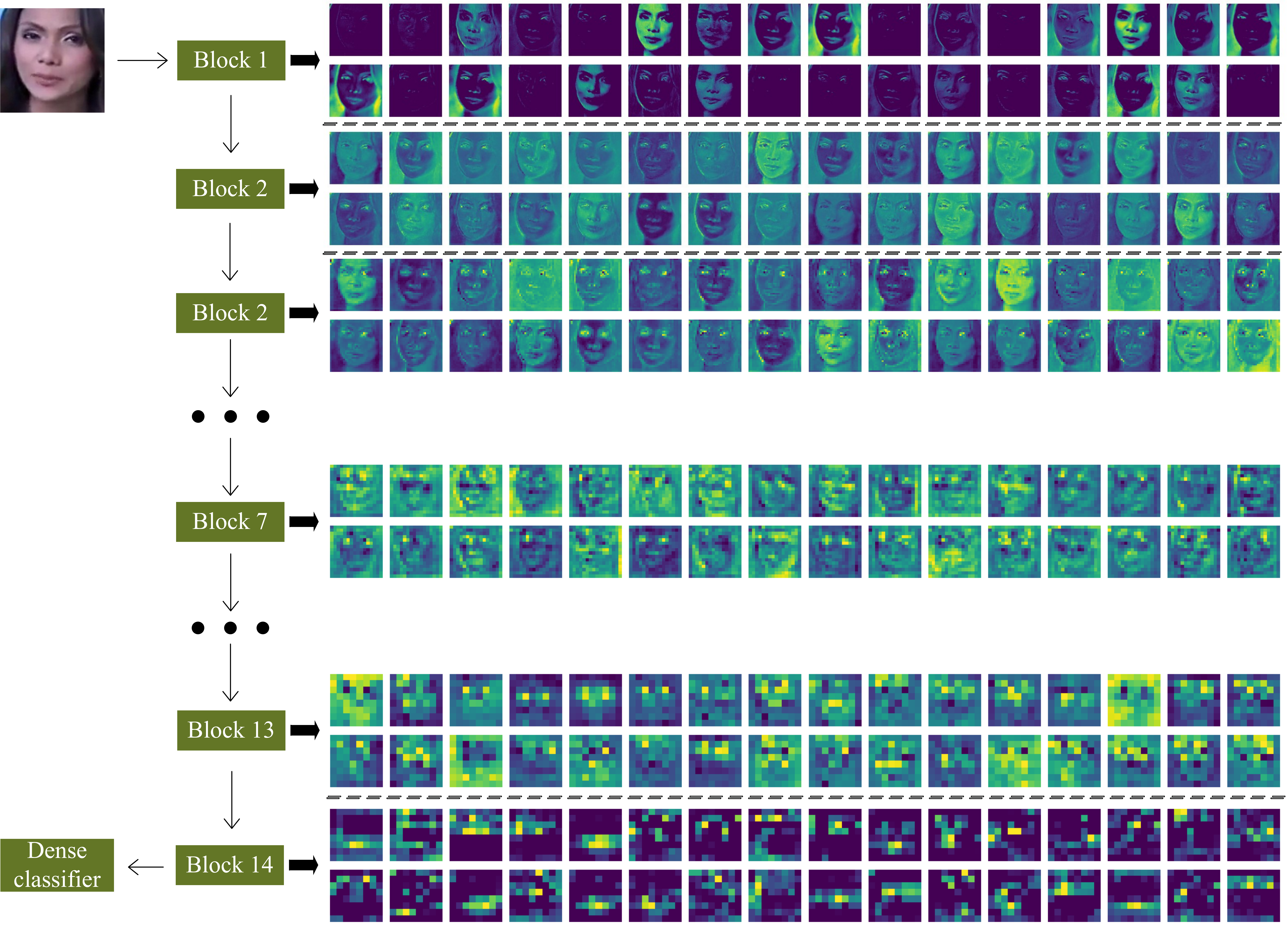}
  \caption{The visualization of feature maps from different blocks in Xception\cite{chollet2017xception} network, where the model was trained for deepfake detection.}
\label{fig:Act_blocks}
\end{figure*}

\begin{figure*}[!htb]
  \centering
  \includegraphics[width=1\linewidth]{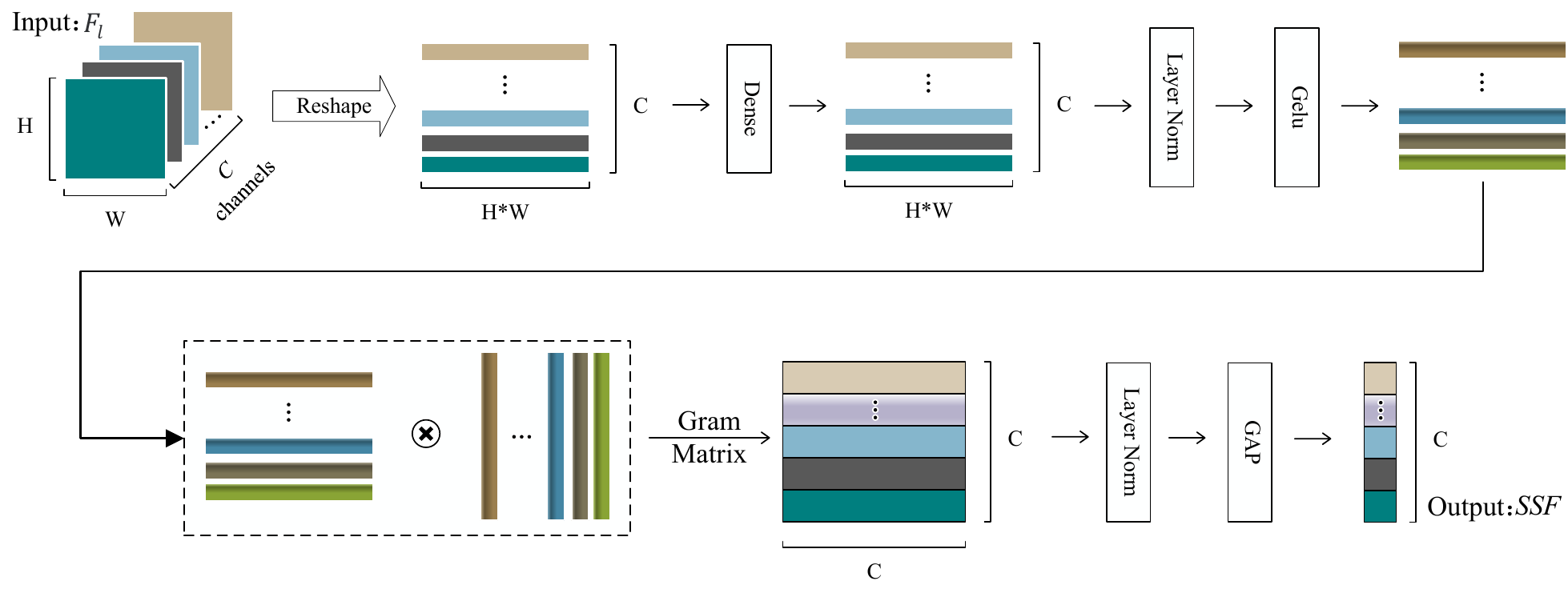}
  \caption{The structure of shallow style representation block (SSRB). This module acts on truncated feature maps $F_{l}$ of the backbone network and produces shallow style features $\textit{SSF}$.}
\label{fig:SSRB}
\end{figure*}

\subsection{Shallow style representation block} \label{sec:SSRB}

The shallow layers of the deep convolutional neural network have smaller receptive fields and tend to distinguish low-level features (e.g., color and texture) locally, while the deeper layers tend to be high-level abstract features. Figure \ref{fig:Act_blocks} shows feature maps at different layers of the trained deepfake detection network. It can be observed that the feature maps output by shallower layers (blocks 1-3) are generally activated on a small scale and identify local regions such as the eyes, nose, and mouth. With the deepening of layers and the increase of receptive fields, the activated regions of feature maps become larger, which means that higher-level semantic concepts are encoded. The feature maps of the last blocks 13-14 present abstract activations rather than local filtering patterns. In deep fake detection tasks, many methods directly connect binary classifier to the last layer of CNN and decide with global features. However, given those common facial organs, there are few differences between forged and real faces in terms of high-level semantic information. Moreover, some manipulation methods only tamper with the specified region of a face. Therefore, the forgery artifacts are usually more discriminative in shallow feature maps. Based on these phenomena, we extract multi-scale low-level style representations through the Gram matrix at the shallow layers of the backbone network, and learn with other modal features, as shown in the top stream of Figure \ref{fig:Framework}.

Gram matrix is usually used to encode style attributes in deep neural networks\cite{gatys2015texture, gatys2016image}, such as shading and texture patterns. Specifically, it calculates the dot product of the vectorized feature maps $F_{i}^{l}$ and $F_{j}^{l}$ of the $l$-th layer:
\begin{equation}
\centering
G_{i j}^{l}=\sum_{k} F_{i k}^{l} F_{j k}^{l}=\left[\begin{array}{ccc}
F_{i 1}^{l}{ }^{T} F_{j 1}^{l} & \cdots & F_{i 1}^{l}{ }^{T} F_{j k}^{l} \\
\vdots & \ddots & \vdots \\
F_{i k}^{l}{ }^{T} F_{j 1}^{l} & \cdots & F_{i k}^{l}{ }^{T} F_{j k}^{l}
\end{array}\right],
\end{equation} where Gram matrix $G_{i j}^{l}$ is the eccentric covariance matrix, i.e., the covariance matrix of the $i$-th and $j$-th feature maps in layer $l$, but without subtracting the mean value. The Gram matrix actually measures the pairwise correlation of two feature maps: the diagonal elements provide the respective responses of each filter to itself, and the remaining elements indicate the degree of correlation between the different filter responses. We argue that such pairwise correlations that emphasize response patterns contribute to capturing forged textures between different feature maps, taking into account those stereotyped visual artifacts\cite{matern2019exploiting} or face warping artifacts\cite{li2018exposing}.

Given the remarkable performance of Xception in deepfake detection\cite{rossler2019faceforensics++}, we employ it as the backbone network, and carefully design the shallow style representation block (SSRB) to extract discriminative local texture features. As shown in Figure \ref{fig:SSRB}, SSRB truncates the feature map $F_{l} \in R^{H \times W \times C}$ of layer $l$ as input, and refines $F_{l}$ as a vectorized feature representation. Considering that the calculated Gram matrix $G_{l} \in R^{C \times C}$ has redundant correlation and is hard to be aligned in multiple semantic layers, we aggregate the cumulative correlation of each feature map (i.e., each row of $G_{l}$) through the global average pooling layer, and finally, obtain the shallow style feature $\textit{SSF} \in R^{C \times 1}$. There are 14 blocks in Xception\cite{chollet2017xception}, which are denoted from $b1$ to $b14$. Specifically, SSRB truncates the feature maps with smaller receptive fields output by blocks 1-3 and obtains $\textit{SSF}_{b1}$, $\textit{SSF}_{b2}$, and $\textit{SSF}_{b3}$, respectively. As shown in Figure \ref{fig:Framework}, the extracted $\textit{SSF}$ features act as multi-scale shallow style representations, given that the feature maps of different blocks correspond to receptive fields of different scales. These shallow style features complement the global semantic feature $\textit{GF}$ output by block 14, encouraging the network to capture more subtle locally correlated texture information in the spatial domain.

\begin{figure*}[!htb]
  \centering
  \includegraphics[width=\linewidth]{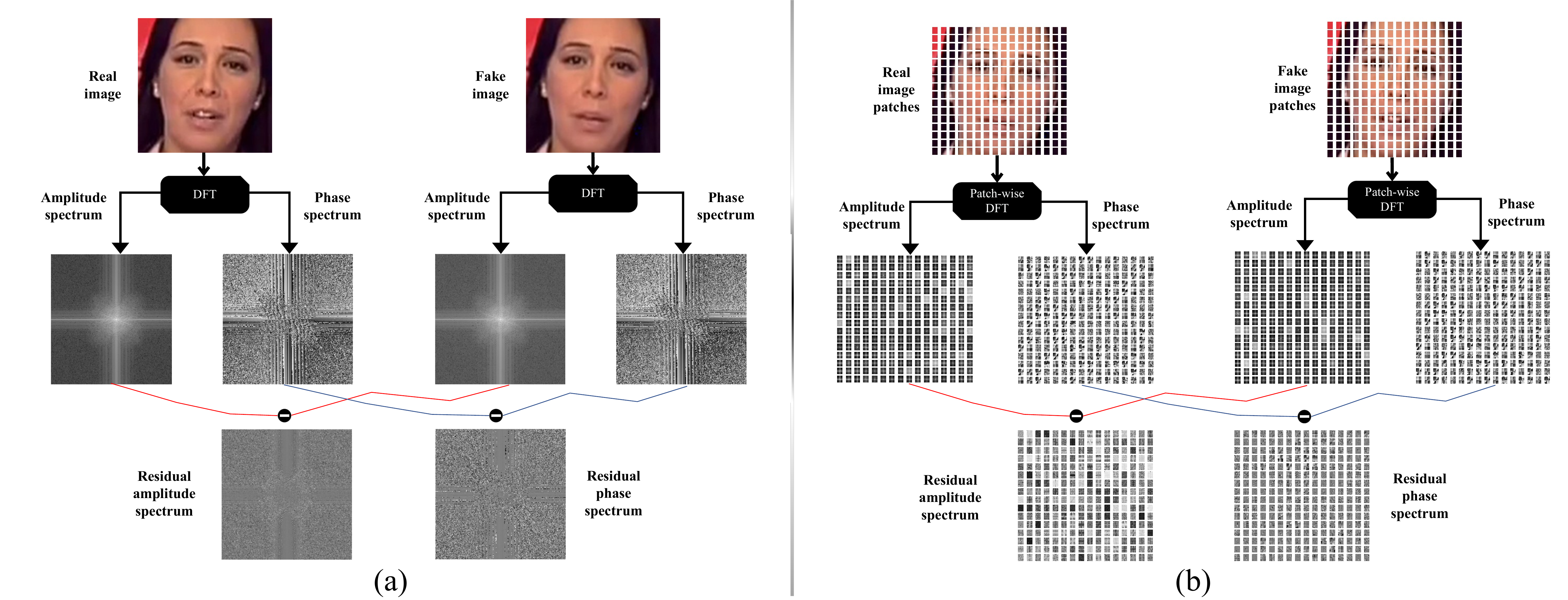}
  \caption{The frequency domain analysis of the (a) entire image and (b) corresponding patches. Residual images between real and forged ones show that the patch-wise amplitude spectrum successfully exposes implicit checkerboard artifacts, and the patch-wise phase spectrum has more differences (brighter color indicates smaller pixel values). The fake image is a NeuralTextures-based forgery face from FF++(LQ)\cite{rossler2019faceforensics++}.}
\label{fig:APS}
\end{figure*}

\subsection{Amplitude and phase anomalies of forged faces} \label{sec:PWAPS}

The up-sampling structure will introduce periodic checkerboard artifacts in the frequency domain. Some methods\cite{frank2020leveraging, durall2019unmasking, zhang2019detecting} directly extract the frequency spectrum from the entire image and train the binary classifier to detect images generated by various GANs. However, in some faces manipulated by computer graphics-based methods and improved deepfake videos\cite{thies2019deferred,li2020celeb}, forged artifacts are insignificant in the global frequency domain of the entire image, as shown in Figure \ref{fig:APS}-(a). In view of the forged cues implied by different frequency bands (especially high-frequency components), other methods\cite{masi2020two, zhou2018learning} extract features through hand-crafted filters as inputs to the model, but with the concomitant consequence that the complete frequency domain is not covered. Further, the phase information of images is lost when only the amplitude spectrum is used for detection\cite{durall2019unmasking, zhang2019detecting}, and vice versa\cite{liu2021spatial}. To address these issues, we propose to extract the patch-wise discriminative frequency representation on amplitude and phase spectrum simultaneously.

As shown in Figure \ref{fig:Framework}, we convert the input image $I$ with the resolution of $H \times W \times C$ to grayscale, i.e., $C$=1. Because the use of RGB images in our experiments significantly increases the dimension of frequency features, but it does not significantly improve performance. Then we split it into $N={HW}/{P^{2}}$ non-overlapping patches, and the size of each patch is $P \times P \times 1$. After that, each patch $f_{i}(x, y)$ is independently performed discrete Fourier transform, which is:
\begin{equation}
\begin{gathered}
F_{i}(u, v)=\frac{1}{P^{2}} \sum_{x=0}^{P-1} \sum_{y=0}^{P-1} f_{i}(x, y) e^{-2 \pi j\left(\frac{u x+v y}{P}\right)}, \\
\text { for } i=1, \ldots, N,
\end{gathered}
\end{equation}
and expansion with Euler's formula can be further expressed as:
\begin{equation}
F_{i}(u, v)=R_{i}(u, v)+j I_{i}(u, v)=\left|F_{i}(u, v)\right| e^{j \varphi_{i}(u, v)}.
\end{equation}
Here, we acquire the amplitude spectrum $AS_{i}$ and phase spectrum $PS_{i}$ for each patch, as follows:
\begin{equation}
\begin{gathered}
AS_{i}=\left|F_{i}(u, v)\right|=\left[R_{i}^{2}(u, v)+I_{i}^{2}(u, v)\right]^{\frac{1}{2}}, \\
PS_{i}=\varphi_{i}(u, v)=\arctan \left[\frac{I_{i}(u, v)}{R_{i}(u, v)}\right], \\
\text { for } i=1, \ldots, N,
\end{gathered}
\end{equation}
and the size of $AS_{i}$ and $PS_{i}$ is also $P \times P$.

The phase spectrum generally holds the structure and position information of the image, and the amplitude spectrum actually contains most texture and shading information. Amplitude and phase information complement the loss of each other, and both play an important role in image perception\cite{morgan1991relative}. A visual comparison of the $AS_{i}$ and $PS_{i}$ of all patches from an original image and its corresponding forged image is provided in Figure \ref{fig:APS}-(b). It is apparent that the $AS_{i}$ and $PS_{i}$ of the subdivided patches are more diversified than the global spectrum $AS$ and $PS$. Pay attention to the residual amplitude spectrum at Figure \ref{fig:APS}-(a), the forged image does not clearly expose frequency artifacts that distinguish it from the real image. Behind this phenomenon, the fake image has only been altered a very limited part of the pristine face, resulting in only slight brightness differences in the global frequency domain. Fortunately, the residual amplitude spectrum in Figure \ref{fig:APS}-(b) exposes many apparent checkerboard artifacts on split patches. This motivated us to explore frequency information on local patches. Further comparing the residual images of the patch-wise amplitude spectrum and the phase spectrum, the pixel difference of the phase spectrum exists in more patches, which indicates that the phase spectrum presents more inconsistencies. 

Based on these observations, we propose the patch-wise amplitude and phase dual attention (PAPDA) module, which explicitly considers the correlation between the amplitude and phase information to extract more discriminative frequency features.

\subsection{Patch-wise amplitude and phase dual attention module}\label{sec:PAPDA}

\begin{figure*}[!htb]
  \centering
  \includegraphics[width=.8\linewidth]{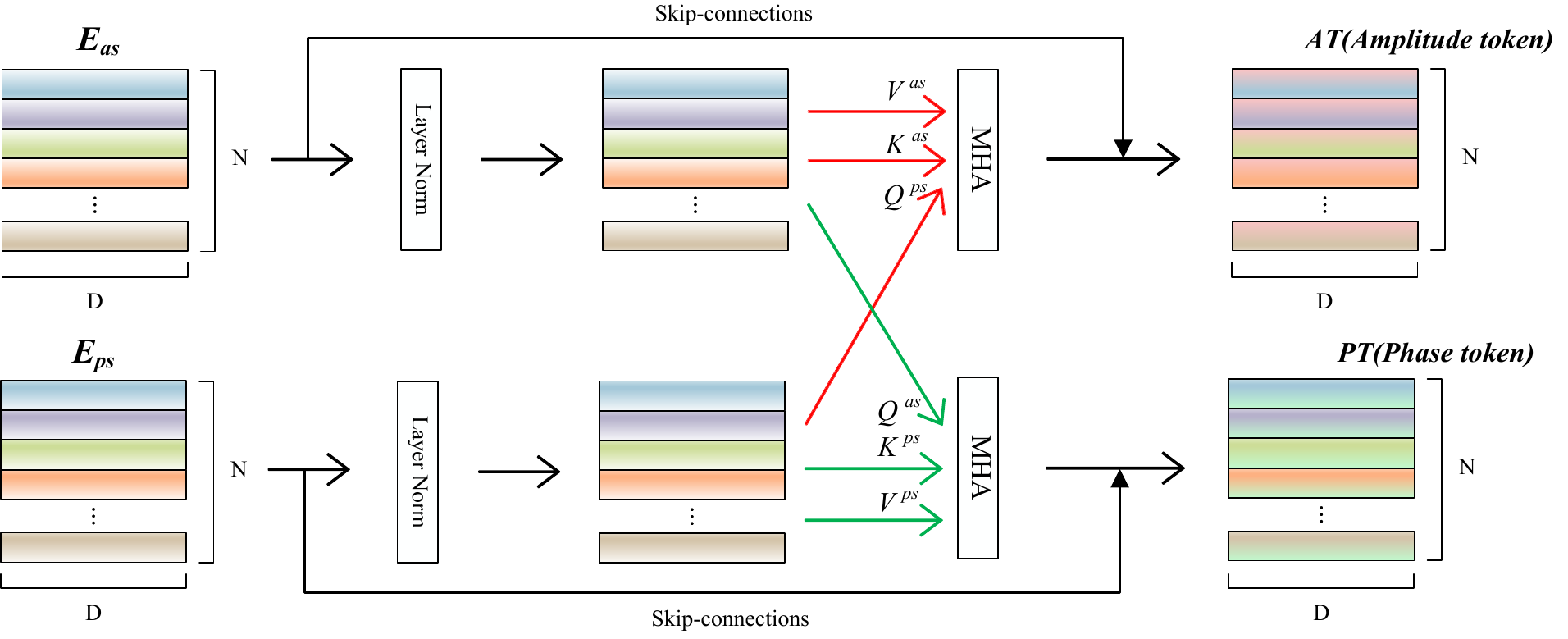}
  \caption{The patch-wise amplitude and phase dual attention (PAPDA) module.}
\label{fig:PAPDAM}
\end{figure*}

\begin{figure*}[htb]
  \centering
  \includegraphics[width=\linewidth]{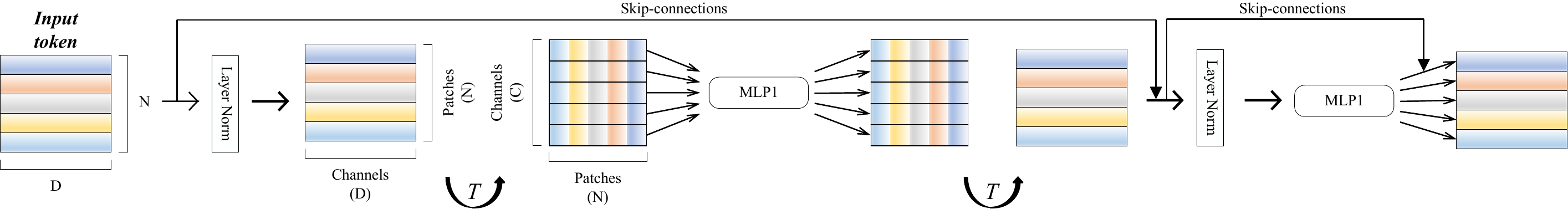}
  \caption{The structure of a Mixer layer\cite{tolstikhin2021mlp}.}
\label{fig:MLP}
\end{figure*}

First, the patch-wise amplitude spectrums $AS_{i}$($i=1,...,N)$ are flattened and projected linearly to a $D$-dimensional embedding $E_{as} \in \boldsymbol{R}^{N \times D}$ through a shared-weight dense layer, which keep the position relationship of image patches in the frequency domain by embedding layer. Here we set $D$ to be the same as the input dimension, i.e., $D=P \times P$. In the same way, the phase spectrums $PS_{i}$ are projected to the phase embedding $E_{ps} \in \boldsymbol{R}^{N \times D}$ through another linear layer. Aims to complement the information of the amplitude and phase fully, we designed the PAPDA module to guide the interactions between $E_{as}$ and $E_{ps}$, as shown in Figure \ref{fig:PAPDAM}.

PAPDA is inspired by the attention module of transformer\cite{vaswani2017attention}, and treats $E_{as}$ or $E_{ps}$ as query($Q$), key($K$), and value($V$). Specifically, in the amplitude attention branch, $K_{as}=V_{as}=LN(E_{as})$ but $Q_{ps}=LN(E_{ps})$, where $LN$ represents layer normalization\cite{ba2016layer}. The output matrix is calculated as:
\begin{equation}
A\left(Q^{ps}, K^{as}, V^{as}\right)=\operatorname{softmax}\left(\frac{Q^{ps}\left(K^{as}\right)^{T}}{\sqrt{D}}\right) V^{a s},
\end{equation}
where $Q^{ps}$ and $K^{as}$ are used to measure the correlation between the amplitude artifacts and the phase differences, such that $Q^{ps}\left(K^{as}\right)^{T}$ as an interactive weight to enhance the amplitude embedding $E_{as}$. We use beneficial multi-head attention\cite{vaswani2017attention}: the attention function is calculated in parallel on $d$-dimensional $Q$, $K$, and $V$ after linear projection by $h$ heads, and then all the outputs are connected and projected back again. Note that $d$ is set to $D/h$ to keep the consistency of dimension and calculation. These can be summarized as follows: 
\begin{equation}
\begin{gathered}
MHA\left(Q^{ps}, K^{as}, V^{as}\right)=\left[A_{1}, A_{2}, \ldots, A_{h}\right] W^{0}, \\
A_{i}=A\left(Q^{ps} W_{i}^{Q}, K^{as} W_{i}^{K}, V^{as} W_{i}^{V}\right),
\end{gathered}
\end{equation}
where $W_{i}^{Q} \in \boldsymbol{R}^{D \times d_{q}}$, $W_{i}^{K} \in \boldsymbol{R}^{D \times d_{k}}$, $W_{i}^{V} \in \boldsymbol{R}^{D \times d_{v}}$ and $W^{O} \in \boldsymbol{R}^{h \times d_{v} \times D}$ are the projection matrices. We use 4 attention heads, i.e., $d_{q}=d_{k}= d_{v}= D_s/h=256/4$. Finally, we obtain the amplitude token $AT$ through the residual connection\cite{he2016deep}:
\begin{equation}
AT=E_{as}+MHA\left(Q^{ps}, K^{as}, V^{as}\right).
\end{equation}
As shown in Figure \ref{fig:PAPDAM}, the phase token $PT$ are obtained through similar computation:
\begin{equation}
PT=E_{ps}+MHA\left(Q^{as}, K^{ps}, V^{ps}\right).
\end{equation}

The spectral representation is not compatible with vanilla CNNs, given that the frequency domain does not match the shift-invariance and local consistency owned by natural images. Unlike other methods\cite{liu2021spatial, qian2020thinking, wang2021m2tr} that convert the extracted frequency representation back to color-space again, we aim at utilizing the frequency information directly: a two-stream MLP-Mixer\cite{tolstikhin2021mlp} takes $AT$ and $PT$ as input, respectively, and capture locally correlated inconsistencies in amplitude and phase between different patches. In specific, amplitude or phase tokens $T \in \boldsymbol{R}^{N \times D}$ are input to a sequence of Mixer layers. As shown in Figure \ref{fig:MLP}, each layer is identical and consists of a token-mixing MLP block and a channel-mixing MLP block. The calculation of the token-mixing MLP block is as follows:
\begin{equation}
U=T+W_{2}\left[\sigma\left(W_{1}\left(L N(T)^{T}\right)\right)\right]^{T},
\end{equation}
where $W_{1}$ and $W_{2}$ are linear operations by the fully connected layer. $LN$ and $\sigma$ represent layer normalization\cite{ba2016layer} and GELU\cite{hendrycks2016gaussian} nonlinear activation, respectively. The mapping $\boldsymbol{R}^{N} \rightarrow \boldsymbol{R}^{N}$ is shared across all patches, mixing the information of different patches by acting on each channel independently. After that, the resulting $U \in \boldsymbol{R}^{N \times D}$ is input to the channel-mixing MLP block:
\begin{equation}
\mathrm{Y}=\mathrm{U}+\mathrm{W}_{4}\left[\sigma\left(\mathrm{W}_{3}(\mathrm{LN}(\mathrm{U}))\right)\right].
\end{equation}
Similarly, $\mathrm{Y} \in \boldsymbol{R}^{N \times D}$ aggregates all channel information of each patch and is input to the next mixer layer.

The channel-mixing block applies the same linear transformation to all channels ($D$-dimensional vectors) in each patch, i.e., parameter sharing, which is similar to $1 \times 1$ convolution. The channel-mixing block extracts discriminative features from local patches, and such local representation is beneficial to capture stereotyped frequency patterns (such as checkerboard artifacts) in different patches. In contrast, the token-mixing block performs the linear transformation on different patches ($N$-dimensional vectors) across each channel, which is analogous to a depthwise convolution with shared weights on the channels. The Mixer layer fits the correlation between patch-wise amplitude and phase spectrums and provides a good way for long-distance interactions. As the branch network of our model, it complements the local structure awareness of CNN. In our experiments, we only use two Mixer layers as a compromise between detection performance and network scale. $AT$ and $PT$ are input into two MLP-Mixer networks. Finally, the amplitude feature ($AF$) and phase feature ($PF$) are output through the global average pooling layer, that is:
\begin{equation}
\begin{aligned}
&AF=GAP(LN(\operatorname{Mixer}(AT))), \\
&PF=GAP(LN(\operatorname{Mixer}(PT))).
\end{aligned}
\end{equation}

\subsection{Multi-modal features contrastive classification}\label{sec:MFCC}

As shown in Figure \ref{fig:Framework}, the output of our hybrid network includes multi-scale shallow style features $\textit{SSF}_{b1}$, $\textit{SSF}_{b2}$, $\textit{SSF}_{b3}$, high-level semantic features $GF$, amplitude features $AF$ and phase features $PF$. These features of multiple modalities and even different scales are concatenated and used for more comprehensively  face forgery detection:
\begin{equation}
F_{MM}=\operatorname{concatenate}(\textit{SSF}_{b1}, \textit{SSF}_{b2}, \textit{SSF}_{b3}, \textit{GF}, \textit{AF}, \textit{PF}).
\end{equation}
Typically, the feature fusion $F_{MM}$ is followed by a fully connected layer to predict binary classes, supervised by the cross-entropy loss. However, forged face detection is a vanilla binary classification problem, and the classification margin learned by the dichotomy model may not be suitable for forgery methods that are not seen in the training, i.e., it suffers from overfitting. Considering the diversity of manipulation methods, disentangling the representation between different data classes will improve our model's generalization ability. Given the performance improvement brought by supervised contrastive learning\cite{khosla2020supervised} in representation learning, we combine it with the cross-entropy loss to make our model not only learns classifier, but also learns more separate representations.

The idea behind supervised contrastive loss\cite{khosla2020supervised} is to pull the samples from the same class together, and push it away from samples of different classes in the normalized embedding space. Unlike the unsupervised contrastive loss, the supervised contrastive loss takes the same class images as positives, and the label information is effectively used to control the distance. We employ it as one of the loss functions to improve the generalizability of our model. As shown in Figure \ref{fig:Framework}: firstly, $F_{MM}$ is encoded as a representation vector ${F}_{E} \in \boldsymbol{R}^{E}$ through a single linear layer; and then $F_{E}$ is projected to $z \in \boldsymbol{R}^{p}$ with another linear layer. The projected output $z$ is normalized to calculate the inner product for measuring the distance in the embedding space. If the mini-batch size is $N$, the supervised contrastive loss is formulated as:
\begin{equation} \label{eq:SC}
L_{SC}=\sum_{i=1}^{N} \frac{-1}{\left|\left\{z_{i}^{0}\right\}\right|} \sum_{z_{p} \in\left\{z_{i}^{0}\right\}} \log \frac{\exp \left(z_{i} \cdot z_{p} / \tau\right)}{\sum_{z_{n} \in\left\{z_{i}^{1}\right\}} \exp \left(z_{i} \cdot z_{n} / \tau\right)},
\end{equation}
where $0$ and $1$ are used to indicate the class of images, i.e., real or fake, and $\tau \in \boldsymbol{R}^{+}$ is a scalar temperature parameter. $\left|\left\{z_{i}^{0}\right\}\right|$ denotes the number of positive samples. The index $i$ is called an anchor for comparison, $p$ is a positive of the same class as the anchor, and $n$ indicates a negative of a different class.

With the supervision of supervised contrastive loss, the encoded representations $F_{E}$ from the same class are encouraged the compactness. Finally, the prediction $\widehat{y}_{i}$ is obtained through $F_{E}$ followed by the fully connected layer, and learned by cross-entropy loss:
\begin{equation}
L_{CE}=-\frac{1}{N} \sum_{i=1}^{N} y_{i} \log \widehat{y}_{i}+\left(1-y_{i}\right) \log \left(1-\widehat{y}_{i}\right).
\end{equation}
Note that we did not adopt the two-stage training\cite{khosla2020supervised}, but let the model learn the feature representation and the classifier simultaneously, because we found that it would take into account both detection and generalization ability. That is:
\begin{equation} \label{eq:loss}
L=\alpha L_{SC}+(1-\alpha) L_{CE},
\end{equation}
where $\alpha$ is the hyperparameter used to control the trade-off between $L_{CE}$ and $L_{SC}$.

\section{Experiments} \label{sec:Exp}
In this section, we first introduce the overall experimental setup and then present extensive experimental results to demonstrate the effectiveness of the proposed method.

\subsection{Experimental setup}

\noindent\textbf{Datasets} To comprehensively evaluate the effectiveness of our method for detecting forged faces, its robustness in different compression scenarios, and its generalization across datasets, we conduct extensive experiments on three large-scale face forgery benchmarks: FaceForensics++(FF++)\cite{rossler2019faceforensics++}, Celeb-DF\cite{li2020celeb}, and DeeperForensics\cite{jiang2020deeperforensics}. FF++ has been widely evaluated by recent detection models, which is composed of 1000 pristine videos as well as 4000 videos forged by four common face manipulation methods: DeepFakes(DF)\cite{DeepFakes}, Face2Face(F2F)\cite{thies2016face2face}, FaceSwap(FS)\cite{FaceSwap}, NeuralTextures\cite{thies2019deferred}. Each video in FF++ comes in three versions according to the compression level: uncompressed c0 (RAW), lightly compressed c23(HQ), and heavily compressed c40 (LQ). We adopt the HQ and more challenging LQ versions since the benchmark already achieved nearly perfect detection performance in the RAW version. Following the official preprocessing ways\cite{rossler2019faceforensics++}, we split 720 videos for training, 140 videos for validation, and the remaining 140 videos for testing. After 30 frames of each video are sampled, we use DLIB\cite{king2009dlib} to extract and align the faces from frames, and adjust facial image size to $256 \times 256$.

\begin{table}[!htb]
\centering
\caption{Specifications of our MC-LCR architecture used in this paper.}
\label{tab:detail}
\resizebox{\linewidth}{!}{%
\begin{tabular}{l|l|l} 
\hline
\textbf{Abbr}    & \textbf{Connotation}                                               & \textbf{Size}                 \\ 
\hline
$\textit{SSF}_{b1}$            & shallow style feature extracted from Xception b1                   & 64                            \\
$\textit{SSF}_{b2}$            & shallow style feature extracted  from Xception b2                  & 128                           \\
$\textit{SSF}_{b3}$           & shallow style feature extracted from Xception b3                   & 256                           \\
GF               & global feature output from the last block of Xception              & 1024                          \\
P; $AS_{i}$; $PS_{i}$      & patch size                                                         & 16x16x1                       \\
N                & number of split patches                                            & 256(HW/P\textsuperscript{2})  \\
D                & frequency embedding dimension                                      & 256                           \\
$E_{as}$; $E_{ps}$; AT; PT & amplitude embedding; phase embedding; amplitude token; phase token & 256x256(NxD)                  \\
W1               & 1st FC  at token-mixing block                                      & 256(N)                        \\
W2               & 2nd FC at token-mixing block                                       & 256(N)                        \\
W3               & 1st FC at channel-mixing block                                     & 256(D)                        \\
W4               & 1nd FC at channel-mixing block                                     & 256(D)                        \\
AF; PF           & amplitude feature; phase feature                                   & 256                           \\
$F_{E}$               & encoding feature before contrastive learning and classification    & 1024                          \\
Z                & projection representation for contrastive learning                 & 128                           \\
\hline
\end{tabular}}
\end{table}

\begin{table}[!htb]
\centering
\caption{Quantitative detection results in terms of ACC(\%) and AUC(\%) on FF++ dataset with high quality(c23 light compression) and low quality(c40 heavy compression) settings. The \textbf{bold} results are the best.}
\label{tab:FF++com}
\resizebox{\linewidth}{!}{%
\begin{tabular}{c|c|c|c|c|c} 
\hline
\multirow{2}{*}{Methods↓} & Qualities→      & \multicolumn{2}{c|}{HQ(c23)}     & \multicolumn{2}{c}{LQ(c40)}        \\ 
\cline{2-6}
                          & Metrics→        & ACC            & AUC             & ACC             & AUC              \\ 
\hline
\multicolumn{2}{c|}{Steg. Features + SVM\cite{fridrich2012rich}}                                                                       & 70.97          & -               & 55.98           & -                \\
\multicolumn{2}{c|}{LD-CNN\cite{cozzolino2017recasting}}                                                                                     & 78.45          & -               & 58.69           & -                \\
\multicolumn{2}{c|}{C-Conv\cite{bayar2016deep}}                                                                                     & 82.97          & -               & 66.84           & -                \\
\multicolumn{2}{c|}{CP-CNN\cite{rahmouni2017distinguishing}}                                                                                     & 79.08          & -               & 61.18           & -                \\
\multicolumn{2}{c|}{MesoNet\cite{afchar2018mesonet}}                                                                                    & 83.10          & -               & 70.47           & -                \\
\multicolumn{2}{c|}{Xception\cite{rossler2019faceforensics++}}                                                                                & 95.73          & 96.30           & 86.86           & 89.30            \\
\multicolumn{2}{c|}{Xception-ELA\cite{gunawan2017development}}                                                                               & 93.86          & 94.80           & 79.63           & 82.90            \\
\multicolumn{2}{c|}{DSP-FWA\cite{li2018exposing}}                                                                                    & -              & 57.50           & -               & 62.30            \\
\multicolumn{2}{c|}{Face X-ray\cite{li2020face}}                                                                                 & -              & 87.40           & -               & 61.60            \\
\multicolumn{2}{c|}{Two-branch\cite{masi2020two}}                                                                                 & 96.43          & 98.70           & 86.34           & 86.59            \\
\multicolumn{2}{c|}{SPSL\cite{liu2021spatial}}                                                                        & 91.5           & 95.32           & 81.57           & 82.82            \\
\multicolumn{2}{c|}{MADD\cite{zhao2021multi}}                                                                  & 97.6           & 99.29           & 88.69           & 90.40            \\
\multicolumn{2}{c|}{F3-Net\cite{qian2020thinking}}                                                                                     & 97.52          & 98.10           & \textbf{90.43 } & \textbf{93.30 }  \\ 
\hline
\multicolumn{2}{c|}{Ours}                                                                                  & \textbf{97.89} & \textbf{99.65 } & 88.07           & 90.28            \\
\hline
\end{tabular}}
\end{table}

\noindent\textbf{Implementation details and hyper-parameters} We use Xception\cite{chollet2017xception} pre-trained on the imagenet as the RGB branch and MLP-Mixer\cite{tolstikhin2021mlp} as the frequency branch. The drop rate of all dropout layers in our network is set to 0.5. The temperature parameter in the Eq. \ref{eq:SC} is set to 0.1, and the $\alpha$ in the Eq. \ref{eq:loss} is set to 0.5. The batch size is set to 16. Note that each batch consists of 8 real and 8 fake facial images, and is guided for contrastive learning. The Adam\cite{loshchilov2017decoupled} algorithm with initial learning rate $1e-3$ and weight decay $1e-4$ is used to optimize the network. The model is trained for 50 epochs. If the validation loss does not decrease after 5 epochs, the learning rate will be reduced to half of the original. Our experiments are performed on the NVIDIA GeForce RTX 2080Ti GPU with 128GB RAM. More model and feature representation specifications are shown in Table \ref{tab:detail}.

\noindent\textbf{Metrics} Following the evaluation metrics commonly used in deepfake detection tasks\cite{rossler2019faceforensics++, li2020celeb}, we report both the frame-level ACC(accuracy) and AUC(Area Under the Curve of ROC) scores to evaluate the detection performance, which facilitate intuitive comparison with other SOTA methods.

\begin{table*}[!htb]
\centering
\caption{Quantitative detection results in terms of ACC(\%) and AUC(\%) on FF++ dataset with four manipulation methods, i.e. DeepFakes(DF), Face2Face(F2F), FaceSwap(FS), NeuralTextures(NT). The \textbf{bold} results are the best.}
\label{tab:FF++four}
\resizebox{.8\linewidth}{!}{%
\begin{tabular}{c|c|c|c|c|c|c|c|c|c} 
\hline
\multirow{2}{*}{Methods↓} & Manipulations(LQ)→ & \multicolumn{2}{c|}{DF}           & \multicolumn{2}{c|}{F2F}          & \multicolumn{2}{c|}{FS}           & \multicolumn{2}{c}{NT}             \\ 
\cline{2-10}
                          & Metrics(\%)→       & ACC             & AUC             & ACC             & AUC             & ACC             & AUC             & ACC             & AUC              \\ 
\hline
\multicolumn{2}{c|}{Steg.Features\cite{fridrich2012rich}}                                                                                      & 73.64           & -               & 73.72           & -               & 68.93           & -               & 63.33           & -                \\
\multicolumn{2}{c|}{LD-CNN\cite{cozzolino2017recasting}}                                                                                             & 85.45           & -               & 67.88           & -               & 73.79           & -               & 78.00           & -                \\
\multicolumn{2}{c|}{C-Conv\cite{bayar2016deep}}                                                                                             & 85.45           & -               & 64.23           & -               & 56.31           & -               & 60.07           & -                \\
\multicolumn{2}{c|}{CP-CNN\cite{rahmouni2017distinguishing}}                                                                                             & 84.55           & -               & 73.72           & -               & 82.52           & -               & 70.67           & -                \\
\multicolumn{2}{c|}{MesoNet\cite{afchar2018mesonet}}                                                                                            & 87.27           & -               & 56.20           & -               & 61.17           & -               & 40.67           & -                \\
\multicolumn{2}{c|}{Xception\cite{rossler2019faceforensics++}}                                                                                           & 95.13           & 99.24           & 87.34           & 93.62           & 92.42           & 97.08           & 77.54           & 84.51            \\
\multicolumn{2}{c|}{SPSL\cite{liu2021spatial}}                                                                                               & 93.48           & 98.50           & 86.02           & 94.62           & 92.26           & 98.10           & 76.78           & 80.49            \\ 
\hline
\multicolumn{2}{c|}{Ours}                                                                                                & \textbf{97.23 } & \textbf{99.53 } & \textbf{91.08 } & \textbf{96.72 } & \textbf{94.44 } & \textbf{98.60 } & \textbf{82.13 } & \textbf{87.39 }  \\
\hline
\end{tabular}}
\end{table*}

\subsection{Evaluation and comparison on FF++}

\noindent\textbf{Different video compression} We first evaluate our method on different compression versions of the FF++ and report the comparison with previous detection methods in the Table \ref{tab:FF++com}. The test results show that the proposed method outperforms previous methods: 1) Obviously, our method is superior to early methods such as Steg.Features\cite{fridrich2012rich}, LD-CNN\cite{cozzolino2017recasting}, C-Conv\cite{bayar2016deep}, CP-CNN\cite{rahmouni2017distinguishing}, and MesoNet\cite{afchar2018mesonet} in terms of ACC. 2) Our results are consistently better than powerful detection methods Xception\cite{rossler2019faceforensics++} and Face X-ray\cite{li2020face} in different quality settings. The performance gains mainly benefit from PAPDA, as shown in Section \ref{sec:PWAPS} and Figure \ref{fig:APS}: although compressed videos lose many shading and frequency information, noticeable artifacts are still exposed in the patch-wise amplitude and phase spectrums. Furthermore, the AUC of our method significantly exceeds Face X-ray by 12.25\% and 28.68\% at the HQ and LQ versions. Face X-ray relies on the discrepancies around blending boundaries, and heavy compression results in weakened blending traces that limit its detection performance.

Even compared with the most recent state-of-the-arts, our method achieves the leadership position on the FF++(HQ) and obtains competitive results on the LQ version. Two-branch\cite{masi2020two} and SPSL\cite{liu2021spatial} also fuse features from spatial and frequency domains to isolate forged faces. Our method outperforms them in almost all metrics, which suggests the effectiveness of the proposed SSRB for extracting shallow style representations. SPSL suppresses high-level semantic information and focuses on local textures by throwing away many convolutional blocks. This straightforward way improves generalizability but limits in-domain data fitting capability, given the reduced parameter space. However, our results with LQ settings are weaker than F3-Net\cite{qian2020thinking}, since F3-Net is targeted at highly compressed videos but lacks generalization ability. Our SSRB extracts style representation, such as shallow texture information, is sensitive to high compression rate, but it can improve generalization performance, as discussed in ablation studies \ref{sec:abaStu}.

\noindent\textbf{Different manipulation methods} 
We further evaluate the proposed framework against different manipulation methods in FF++. In such a case, the models are trained and tested exactly on the LQ version for each manipulation method. The test results listed in Table \ref{tab:FF++four} prove the effectiveness of our method in detecting various manipulated faces. The SPSL\cite{liu2021spatial} makes use of the global phase differences produced by the up-sampling steps. But some local manipulations, e.g., F2F\cite{thies2016face2face} and NT\cite{thies2019deferred}, would not have noticeable differences in the global phase spectrum. To address this issue, our PAPDA captures more significant amplitude artifacts and phase differences in local patches, which further boosts the performance. It is worth mentioning that NeuralTextures(NT) is the most challenging, which only modifies the lip region pixels corresponding to the facial expressions, resulting in subtle forgery artifacts in the RGB space and frequency spectrum. Compared with the baseline Xception\cite{rossler2019faceforensics++}, our method makes a remarkable improvement in ACC of 4.6\%. Our performance improvement is mainly attributed to SSRB and PAPDA capturing forgery traces in spatial and frequency domains and concentrating on locally exposed artifacts.

\subsection{Generalization evaluation on unseen datasets}

\begin{table}[!htb]
\centering
\caption{Cross-dataset evaluation from FF++ to Celeb-DF(AUC \%). Our method achieves better generalization performance than most existing detection methods, while maintains state-of-the-art AUC on in-dataset(FF++ DeepFakes) evaluation.}
\label{tab:gen}
\resizebox{.8\linewidth}{!}{%
\begin{tabular}{c|c|c} 
\hline
Models          & FF++            & Celeb-DF         \\ 
\hline
Two-stream\cite{zhou2017two}      & 70.10           & 53.80            \\
Meso4\cite{afchar2018mesonet}           & 84.70           & 54.80            \\
MesoInception4\cite{afchar2018mesonet}  & 83.00           & 53.60            \\
FWA\cite{li2018exposing}             & 80.10           & 56.90            \\
Xception-raw\cite{rossler2019faceforensics++}    & 99.70           & 48.20            \\
Xception-c23\cite{rossler2019faceforensics++}    & 99.70           & 65.30            \\
Xception-c40\cite{rossler2019faceforensics++}    & 95.50           & 65.50            \\
Multi-task\cite{nguyen2019multi}      & 76.30           & 54.30            \\
Capsule\cite{nguyen2019capsule}         & 96.60           & 57.50            \\
DSP-FWA\cite{li2018exposing}          & 93.00           & 64.60            \\
Two-Branch\cite{masi2020two}      & 93.18           & 73.41            \\
F3-Net\cite{qian2020thinking}          & 98.10           & 65.17            \\
SMIL\cite{li2020sharp}            & 96.80           & 56.30            \\
EfficientNet-B4\cite{tan2019efficientnet} & 99.70           & 64.29            \\
MADD\cite{zhao2021multi}            & 99.80           & 67.44            \\
SPSL\cite{liu2021spatial}            & 96.91           & \textbf{76.88 }  \\
M2TR\cite{wang2021m2tr}            & 99.50           & 65.70            \\ 
\hline
Ours            & \textbf{99.84} & 71.61            \\
\hline
\end{tabular}}
\end{table}

\begin{table}[!htb]
\centering
\caption{Cross-dataset evaluation from FF++ to DeeperForensics(AUC \%). Our method achieves best generalization performance than compared detection methods, while maintains state-of-the-art AUC on in-dataset(FF++ HQ) evaluation.}
\label{tab:genDFo}
\resizebox{\linewidth}{!}{%
\begin{tabular}{c|c|c} 
\hline
Models           & FF++(HQ) & DeeperForensics  \\ 
\hline
MesoNet\cite{afchar2018mesonet}          & 86.40     & 57.46             \\
Xception\cite{rossler2019faceforensics++}      & 97.73     & 69.98             \\
Durall et.al\cite{durall2019unmasking}     & 85.44     & 52.89             \\
Ensemble of CNNs\cite{bonettini2021video} & 94.44     & 73.45             \\
Face X-ray\cite{li2020face}       & 87.40     & 72.30             \\ 
\hline
MC-LCR(ours)     & \textbf{99.65}     & \textbf{74.58}             \\
\hline
\end{tabular}}
\end{table}

To evaluate the generalization ability of the proposed model, we also conduct the cross-dataset experiment, i.e., the model is trained on FF++ with four manipulations while tested on Celeb-DF\cite{li2020celeb} and DeeperForensics\cite{jiang2020deeperforensics}, respectively. Celeb-DF has significantly reduced visual artifacts with improved synthesis algorithms than FF++. DeeperForensics applies image degradations to the real videos of FF++ and varies in scenarios such as poses and illumination conditions. 

We follow the evaluation settings\cite{li2020celeb} of Celeb-DF to split testing set and report the frame-level AUC scores in the Table \ref{tab:gen}. The results show that our method achieves better generalization than most existing methods while still keeping almost the top AUC in FF++ DeepFakes. The SSRB and $L_{SC}$ significantly improve the transferability of our model, which is discussed in the ablation studies \ref{sec:abaStu}. Two-branch\cite{masi2020two} and SPSL\cite{liu2021spatial} generalize to Celeb-DF better than ours, at the expense of the in-dataset AUC scores are far behind ours. How to balance bias and variance has always been a challenge, and our method also has limitations here. In the evaluation of generalizing to DeeperForensics, we follow \cite{jiang2020deeperforensics} to split 20\% videos in each sub-database as the testing set and report the frame-level AUC scores in Table \ref{tab:genDFo}. The proposed MC-LCR achieves the best performance in both FF++(HQ) and DeepForensics. MC-LCR improves roughly 5\% AUC to the baseline Xception, which demonstrates the robustness of our method for generalizing to real-world perturbations.

\subsection{Evaluation of efficiency}

Considering MC-LCR introduced multiple components for contrastive classification learning in spatial and frequency domains simultaneously, we also compare the efficiency with other models to measure the consumption of resources and computation. Table \ref{tab:effi} reports the number of parameters(Param), floating-point operations(Flops), and total read-write memory cost(Total MemR+W) of the proposed MC-LCR and other Deepfake detection methods. The comparison models include the baseline networks, VGG19, EfficientNet-B4, ResNet50, and Xception, which are widely adopted as backbone networks for various deepfake detection methods. We also reproduced the state-of-the-art DSP-FWA and MADD methods with officially released models. All models are evaluated under the same environment, and the input size of the image is $256 \times 256 \times 3$.

\begin{table}[!htb]
\centering
\caption{Efficiency comparison among several backbone networks and deepfake detection methods.}
\label{tab:effi}
\resizebox{\linewidth}{!}{%
\begin{tabular}{c|ccc} 
\hline
Model                    & Param & GFLOPs & MemR+W  \\ 
\hline
VGG19                    & 20.6M & 25.6 G & 860MB   \\
EfficientNet-B4          & 19.5M & 4.0G   & 222MB   \\
ResNet50                 & 25.7M & 10.9G  & 376MB   \\
Xception                 & 22.9M & 11.9G  & 405MB   \\
DSP-FWA\cite{li2018exposing}(ResNet101-based) & 42.6M & 10.8G  & 487MB   \\
MADD\cite{zhao2021multi}(Xception-based)     & 49.9M & 14.5G  & 470MB   \\ 
\hline
MC-LCR(Xception-based)   & 30.5M & 13.4G  & 446MB   \\
\hline
\end{tabular}}
\end{table}

As shown in Table \ref{tab:effi}, baseline classification networks have fewer parameters, Flops, and memory cost, given that there are no modules introduced or framework redesigned specifically for deepfake detection, which also limits the detection performance of these naive CNN. Compared with the baseline network Xception, the proposed MC-LCR increases the network parameters by 7.6M but is still far less than DSP-FWA and MADD. The additional parameters are mainly derived from the introduced SSRB, PAPDA, and MLP-Mixer modules. However, the size of Flops and Total MemR+W of MC-LCR does not increase significantly to Xception and is much smaller than VGG19, which indicates that the proposed method is acceptable in terms of computational cost and memory overhead. It is worth mentioning that MC-LCR has advantages in both time and space complexity compared with the MADD method that adopted Xception as the backbone network, and achieved better performance and stronger generalization in previous comparisons.

\section{Ablation Studies}\label{sec:abaStu}
In this section, we perform ablation studies to demonstrate the effectiveness of each proposed module. The SSRB is designed to extract the style representation from shallow layers, given that the low-level texture artifacts introduced by different manipulation methods are more discriminative and generalizable in smaller receptive fields. To dig out forgery clues in the frequency domain and against local tampering methods, PAPDA further interactively captures implicated checkerboard artifacts and differences in the patch-wise amplitude and phase spectrum, which complement the RGB branch. After these locally correlated features are extracted from both spatial and frequency domains, the supervised contrastive loss is combined with cross-entropy loss to learn more separate encoding representations, improving generalization ability. We remove each component individually from our framework and keep other settings the same, then quantitatively evaluate these variants: 1) the baseline model(Xception) w/o SSRB, PAPDA, and SC loss $L_{SC}$, 2) our method w/o SSRB, 3) our method w/o PAPDA, 4) our method w/o SC loss, i.e., $\alpha$ in the Eq. \ref{eq:loss} is set to $0$.

\subsection{Effectiveness to in-dataset detection}

\begin{table}[!htb]
\centering
\caption{Ablation results on FF++(HQ) and FF++(LQ) of different variants in the proposed MC-LCR.}
\label{tab:abaIN}
\resizebox{\linewidth}{!}{%
\begin{tabular}{c|c|c|c|c|c} 
\hline
\multicolumn{2}{c|}{\begin{tabular}[c]{@{}c@{}}Datasets→\end{tabular}} & \multicolumn{2}{c|}{FF++(HQ)} & \multicolumn{2}{c}{FF++(LQ)}  \\ 
\hline
\multicolumn{2}{c|}{Variants↓ Metrics→}                                  & ACC   & AUC                   & ACC   & AUC                   \\ 
\hline
1 & Baseline                                                                  & 95.79 & 97.73                 & 86.84 & 89.38                 \\
2 & w/o SSRB                                                                  & 96.54 & 98.59                 & 86.93 & 89.44                 \\
3 & w/o PAPDA                                                                 & 96.26 & 98.61                 & 86.05 & 88.83                 \\
4 & w/o SC loss                                                               & 96.91 & 99.38                 & 87.53 & 89.65                 \\ 
\hline
5 & Ours(MC-LCR)                                                                & 97.21 & 99.62                 & 88.07 & 90.28                 \\
\hline
\end{tabular}}
\end{table}

As listed in Table \ref{tab:abaIN}, we conduct experiments on HQ and challenging LQ versions of FF++ to analyze the improved performance of each module in our MC-LCR. The comparison results verify that both SSRB and PAPDA effectively improve the detection performance. Furthermore, variants 2 and 3 show that PAPDA makes more detection performance improvement than SSRB, especially in the LQ version. This is expected, as many existing works\cite{li2021frequency, qian2020thinking, masi2020two} have observed that forgery cues in the frequency domain are generally more robust to compression. Moreover, variant 4 consistently outperforms variants 2 and 3 in terms of ACC and AUC scores, indicating that local features extracted from different domains by SSRB and PAPDA complement each other.

\subsection{Effectiveness to cross-dataset detection}

\begin{table}[!htb]
\centering
\caption{Ablation results of different variants generalized to Celeb-DF(AUC \%).}
\label{tab:abaCR}
\resizebox{.6\linewidth}{!}{%
\begin{tabular}{c|c|c} 
\hline
\multicolumn{2}{c|}{Variants} & Celeb-DF  \\ 
\hline
1 & Baseline                  & 65.29          \\
2 & w/o SSRB                  & 65.41          \\
3 & w/o PAPDA                 & 69.27          \\
4 & w/o SC loss               & 66.76          \\ 
\hline
5 & Ours(MC-LCR)                      & 71.61          \\
\hline
\end{tabular}}
\end{table}

To illustrate the improved generalizability of our framework, Table \ref{tab:abaCR} presents the cross-datasets ablation results. We can observe that the AUC score of our model without SSRB(variant 2) generalized to Celeb-DF is significantly reduced by 6.2\%. This demonstrates that the additionally extracted shallow style representations are more generalized across different forged faces. Plugging SC Loss into training indeed contributes to improving the transferability, which introduced a wider separation margin for multi-modal fusion features between real and fake faces. The AUC score also decreased slightly with the removal of PAPDA(variant 3), although not as severe as removing SSRB and SC loss.

\subsection{Effect of low-level features in different blocks}
\begin{figure}[htb!]
  \centering
  \includegraphics[width=\linewidth]{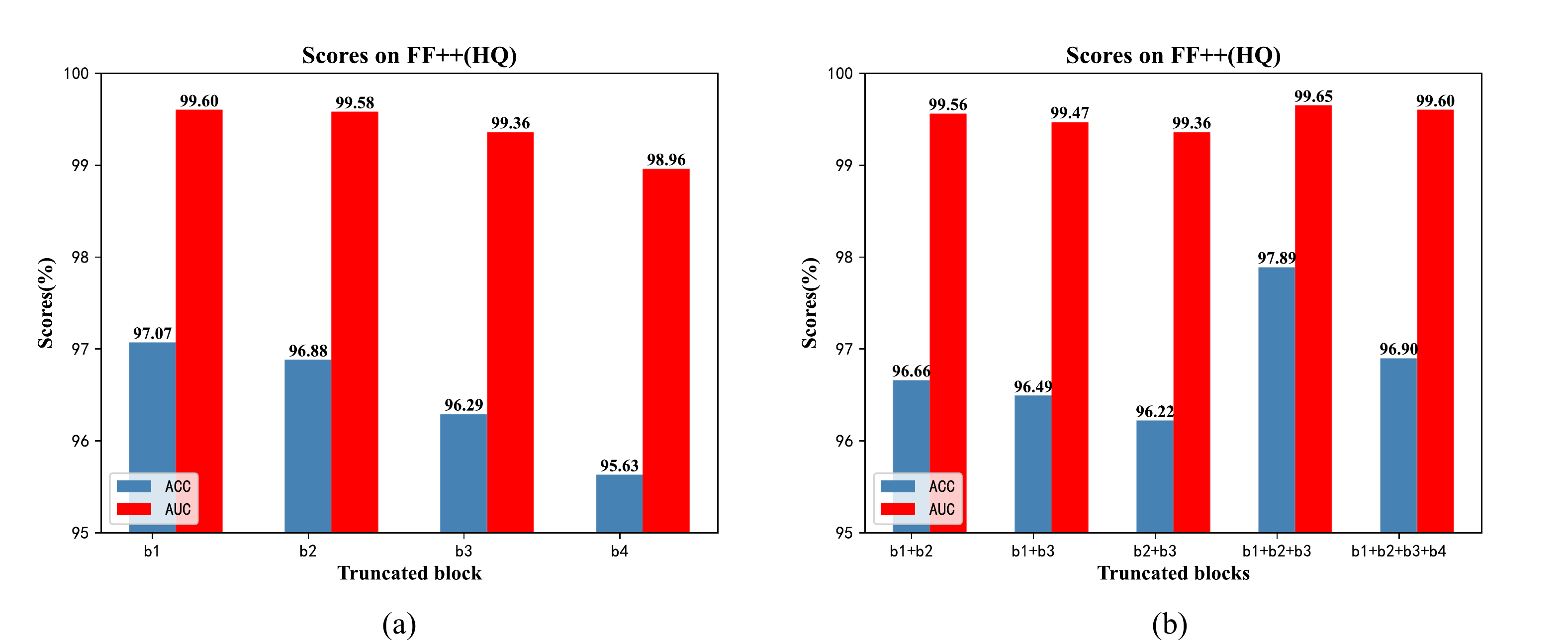}
  \caption{Detection results with features truncated at different blocks of Xception.}
\label{fig:comBlocks}
\end{figure}

Inspired by forgery artifacts of some local face manipulations that are usually more distinct in low-level information, the proposed SSRB truncates the feature maps at the shallow layers b1, b2, and b3 of the backbone network. To illustrate the discriminativeness of low-level local features and the action scope of SSRB, we test SSRB to truncate different blocks and compare detection results in the FF++ dataset. The comparison in Figure\ref{fig:comBlocks}-(a) shows that intercepting local features in shallower layers of the network can achieve better detection performance. Figure\ref{fig:comBlocks}-(b) demonstrates the rationality of SSRB acting on blocks b1, b2, and b3.

\subsection{Effect of different backbones}

\begin{table}[!htb]
\centering
\caption{The comparison of different backbones with the proposed MC-LCR on FF++(LQ) and cross-dataset evaluation on Celeb-DF.}
\label{tab:abaBB}
\resizebox{\linewidth}{!}{%
\begin{tabular}{ccccl} 
\hline
Evaluations→                                & \multicolumn{2}{c}{FF++}                                                & \multicolumn{2}{c}{Celeb-DF}                         \\ 
\cline{2-5}
Backbones↓                                  & ACC                                & AUC                                & ACC            & AUC                                 \\ 
\hline
ResNet50                                    & 83.08                              & 86.78                              & \textbf{66.48} & \textbf{69.99}                      \\
MC-LCR(ResNet50)                            & \textbf{83.51}                     & \textbf{88.19}                     & 64.60          & 68.61                               \\ 
\hline
EfficientNet-B0                             & 83.50                              & 87.51                              & 61.52          & 63.31                               \\
MC-LCR(EfficientNet-B0)                     & \textbf{85.51}                     & \textbf{89.08}                     & \textbf{64.65} & \textbf{66.35}                      \\ 
\hline
EfficientNet-B4                             & 86.03                              & 88.50                              & 61.24          & \multicolumn{1}{c}{63.18}           \\
\multicolumn{1}{l}{MC-LCR(EfficientNet-B4)} & \multicolumn{1}{l}{\textbf{86.48}} & \multicolumn{1}{l}{\textbf{89.68}} & \textbf{62.43} & \multicolumn{1}{c}{\textbf{64.85}}  \\
\hline
\end{tabular}}
\end{table}

Given that the proposed MC-LCR in this paper is mainly based on Xception for experiments and evaluations, we also adopt different networks as the backbone to verify the universality of SSRB, PAPDA, and contrastive classification learning. Three backbones commonly employed in the deepfake detection model, ResNet50, EfficientNet-B0, and EfficientNet-B4, are used for evaluation, where SSRB acts on the shallower three layers. As shown in Table \ref{tab:abaBB}, except that the generalization performance is not improved on ResNet50, other comparison results demonstrate that the proposed method is generally applicable to various backbones to improve performance. We speculate that ResNet50 is larger and deeper than other networks, and such large parameter space leads to the overfitting of the model.

\section{Visualization}

In this section, we make many visualizations to illustrate the effectiveness of our methods, which provide new insights into the success of the proposed MC-LCR.

\subsection{Feature maps} 

\begin{figure}[htb!]
  \centering
  \includegraphics[width=\linewidth]{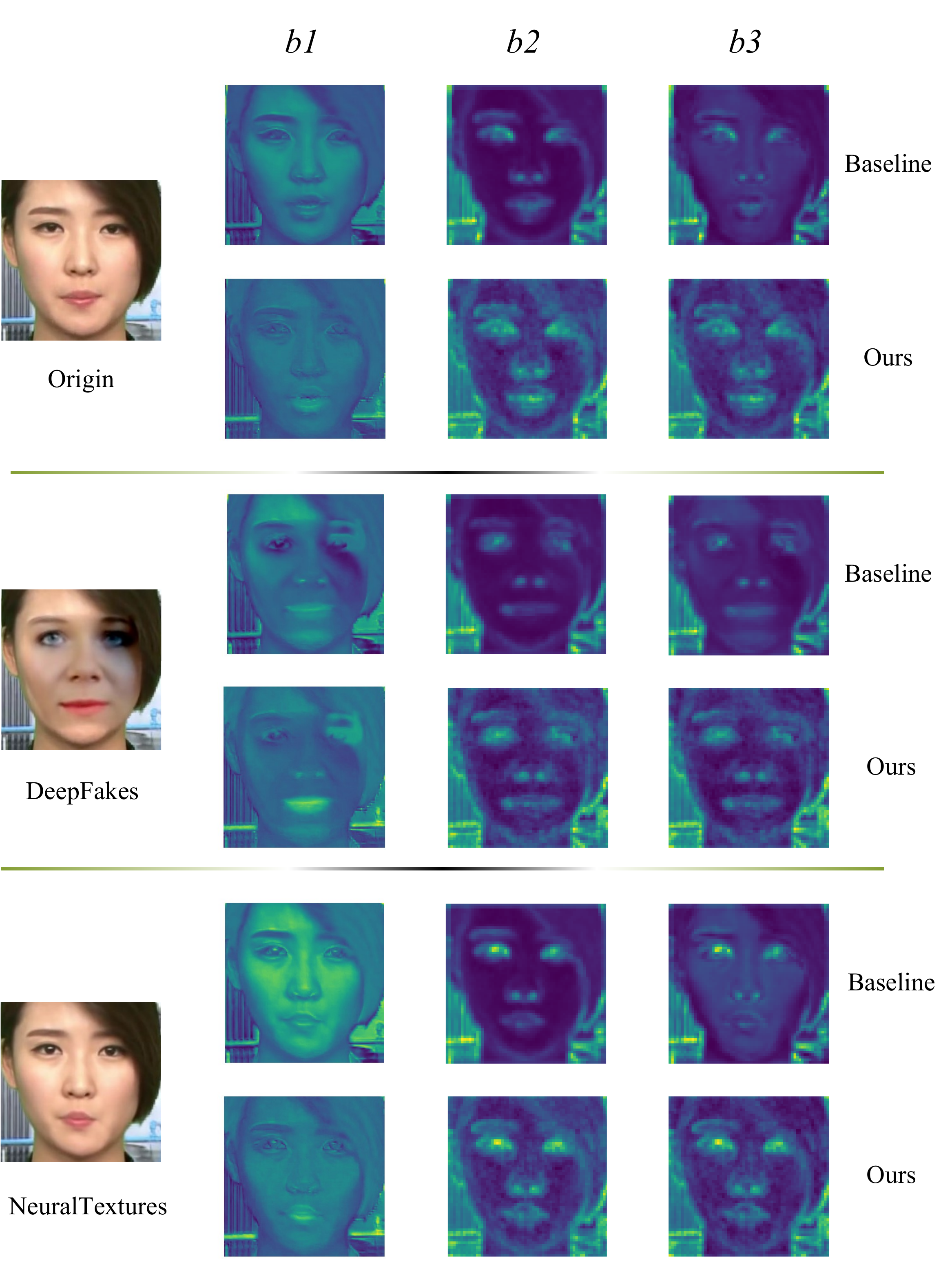}
  \caption{The visualization of feature maps with baseline and our method, where $b1$, $b2$, and $b3$ are extracted from the activation layer of block 1, block 2, and block 3, respectively. Each map is the sum of all feature maps from the corresponding block.}
\label{fig:fm}
\end{figure}

To better understand the effectiveness of our methods, we visualize the shallow feature maps extracted from blocks 1-3 of baseline(Xception) and our MC-LCR, respectively. As presented in Figure \ref{fig:fm}: (1) Almost all feature maps of baseline tend to have similar responses in different facial regions, and our model varies with manipulation methods. There are visible abnormal shadows around the eye and eyebrow of the DeepFakes-based image, and our model makes much higher activation values in this region than the baseline. (2) Our model is more sharp-sighted to subtle forgery artifacts. Note that the NeuralTextures-based image just slightly tamper the mouth, resulting in the baseline, and even human eyes fail to notice. However, our model strongly activates the tampered mouth as early as the shallow layer (b1) of the network. These phenomena are derived from the proposed SSRB and PAPDA, the former measures the pairwise correlation of feature maps, and the latter guides the interaction between the local amplitude and phase spectrums. The feature map visualization illustrates that our methods significantly emphasize the forged region through local correlation, and effectively tackle the problem that the subtle artifacts of small-scale tampering faces are difficult to be captured.

\subsection{Responses of patches}

\begin{figure*}[htb!]
  \centering
  \includegraphics[width=.9\linewidth]{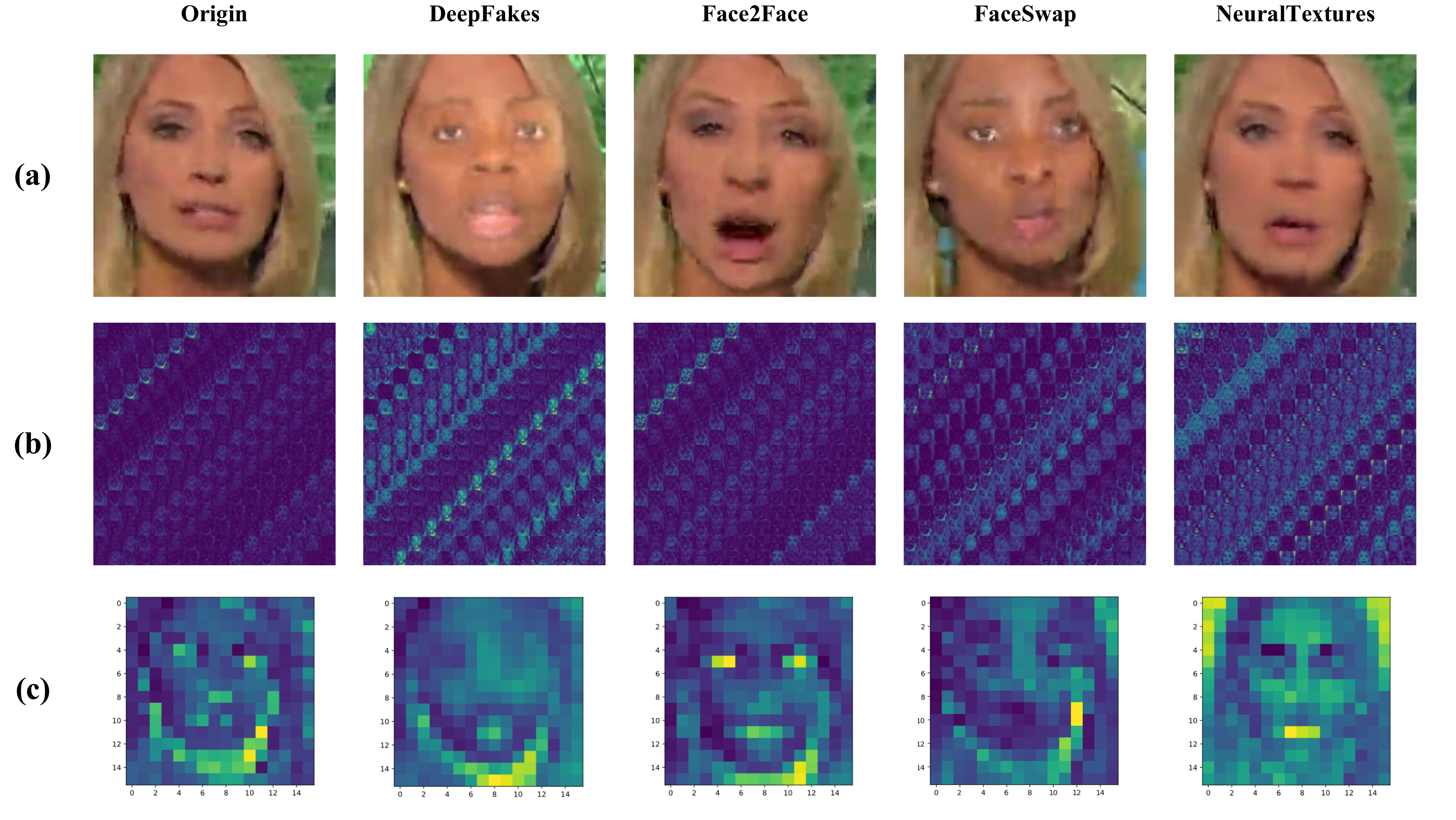}
  \caption{The visualization of patches responses for different manipulations: (a) the input facial image, (b) the response maps of each patch, (c) the average response map of all patches.}
\label{fig:patch_response}
\end{figure*}

In the frequency branch of MC-LCR, the facial image is split into N(256) patches of size $P \times P(16 \times 16)$, which are then transformed to the frequency domain. To visualize the responses of patches to the detected face, we input faces manipulated by different methods into the corresponding trained networks, then extract the frequency embedding and remap its feature values to the input face, as shown in Figure \ref{fig:patch_response}. The response maps of all patches are shown in Figure \ref{fig:patch_response}-(b). Patches can be remapped to the input face given that the spectrum embedding was linearly projected through the fully connected layer, while the diagonal pattern presented by all patches is the embodiment of position encoding. Furthermore, Figure \ref{fig:patch_response}-(c) presents the average response map of all patches. It can be observed that: 1) DeepFakes and FaceSwap are mainly swapped faces, where patches are saliently responded to the face boundaries; 2) Face2Face is an expression manipulation method that controls the facial features, and the responses map successfully focus on the manipulated regions such as eyes and mouth; 3) In particular, NeuralTextures only tampered with the lip, where corresponding tampered patches are strongly activated. These response maps illustrate the effectiveness of the proposed method to capture forgery artifacts in manipulated image patches.

\subsection{Heatmap of feature intensities}

\begin{figure}[htb!]
  \centering
  \includegraphics[width=.9\linewidth]{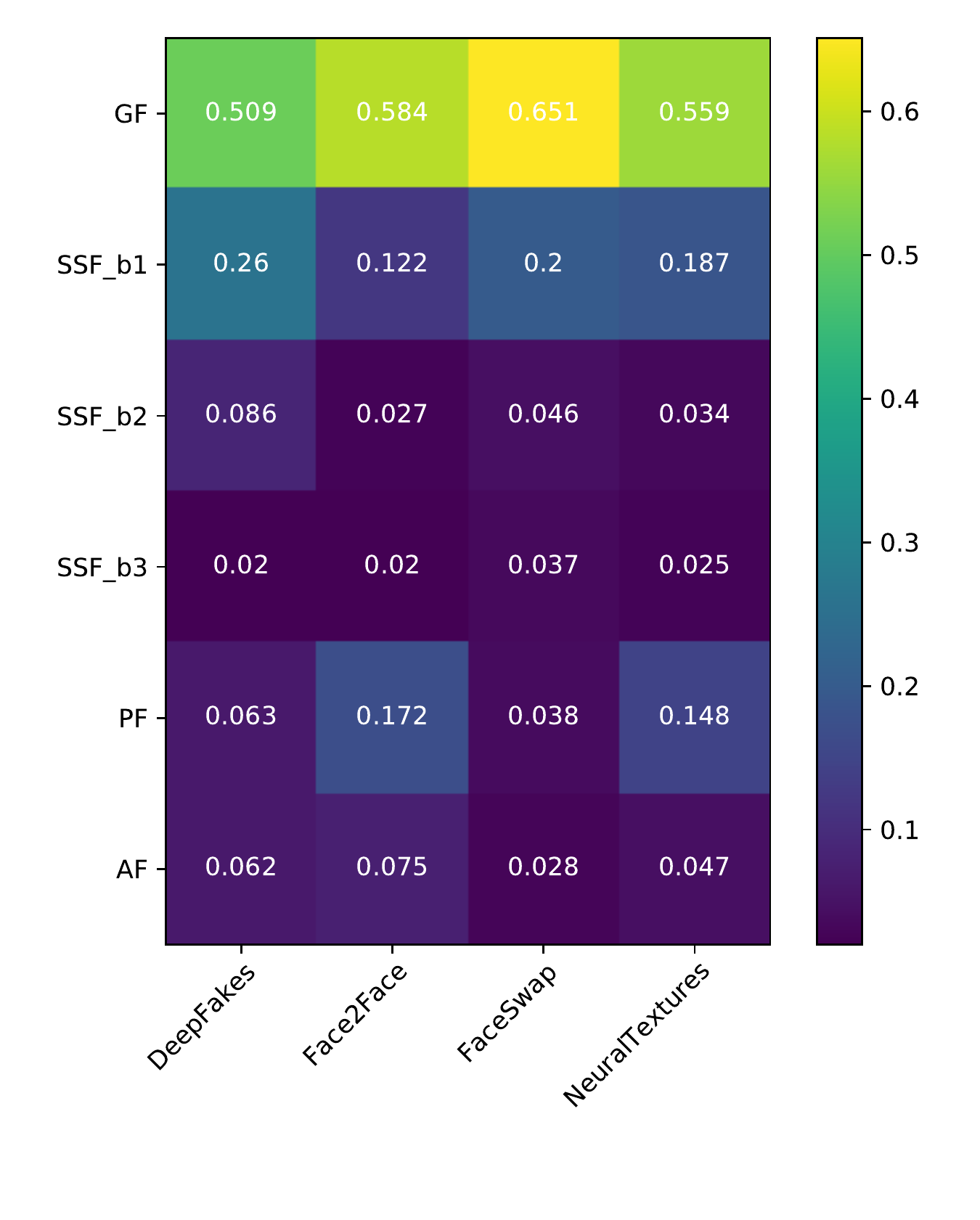}
  \caption{The heatmap of feature activation in the trained MC-LCR model for face forgery detection.}
\label{fig:heatmap}
\end{figure}

The ablation studies in Section \ref{sec:abaStu} demonstrate the effectiveness of different modules in MC-LCR. To further clarify the influence or contribution of features from different components for detection, Figure \ref{fig:heatmap} presents the heatmap of feature intensities, which is derived from the average activation ratio of different features extracted from all testing faces. The global feature $GF$ extracted from the backbone network has the most prominent intensity, especially for faces manipulated by FaceSwap\cite{FaceSwap}. FaceSwap, as a manipulation method based on computer graphics, generally synthesis low-quality faces with salient forged artifacts on the entire image. The activation degree of heat in shallow style features from MC-LCR is ranked as $\textit{SSF}_{b1} > \textit{SSF}_{b2} > \textit{SSF}_{b3}$, which again verifies that the network tends to low-level style features in shallower layers for face forgery detection. As for faces locally manipulated by Face2Face and NeuralTextures, local frequency features $AF$ and $PF$ have a larger proportion than the other two face-swapping methods. This indicates that MC-LCR indeed resorts to capturing more significant amplitude artifacts and phase differences in local frequency patches.

\subsection{Learned encoding representations} 

\begin{figure*}[htb!]
  \centering
  \includegraphics[width=\linewidth]{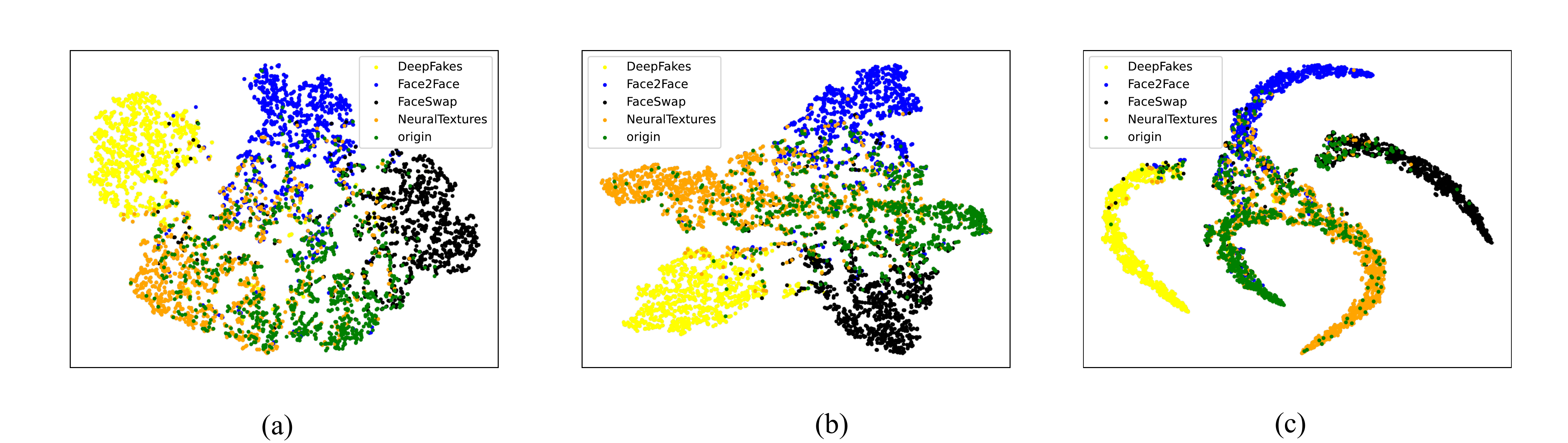}
  \caption{The visualization of t-SNE\cite{van2008visualizing} feature spaces of the (a) baseline, (b) our method w/o SC loss (i.e., variant4), (c) ours(MC-LCR). Green dots represent real faces, and other colors represent the different manipulation faces on FF++(LQ).}
\label{fig:tsne}
\end{figure*}

Furthermore, we also visualize the t-SNE\cite{van2008visualizing} feature spaces of different data in FF++(LQ) to thoroughly explore the influence of our components on the distribution of learned representations $F_{E}$. We can observe from Figure \ref{fig:tsne}: (a) The features extracted from Xception are compactly gathered in the t-SNE embedding space, which limits the discrimination of the four forged faces against real faces. In particular, the features of NT fake faces and real faces are compacted together, because this method only performed small-scale manipulation. (b) After adding the local features extracted from the spatial and frequency domain by SSRB and PAPDA, the distribution of the learned fusion representations has changed. Although some NT fake faces are still confused with real faces, more manipulated faces tend to be farther away from real faces and other categories. These distribution changes prove that the local artifacts captured by our method in different domains help distinguish fake faces from real ones. (c) Further combined with SC loss to training, the representations of the same class are pulled together, while the distances between different classes are significantly boosted. These wider separation boundaries provide new insights into the success of the proposed MC-LCR.

\subsection{Failure Case}

\begin{figure*}[htb!]
  \centering
  \includegraphics[width=\linewidth]{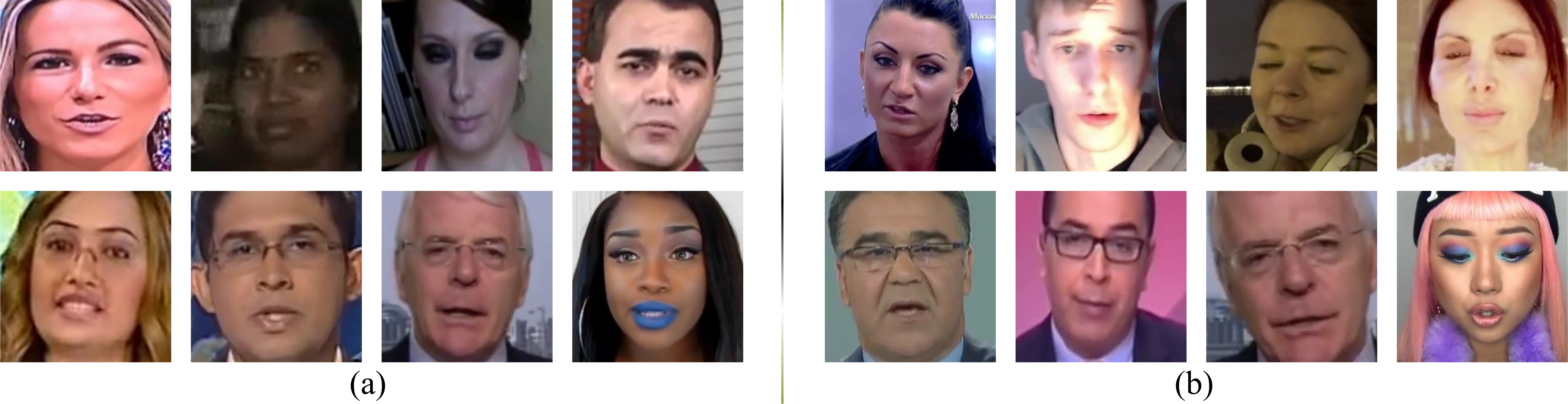}
  \caption{Representative failure cases of MC-LCR: (a) false negatives that are misclassified as real, (b) false positives that are misclassified as fake.}
\label{fig:fc}
\end{figure*}

Although MC-LCR achieved state-of-the-art face forgery detection performance, it is necessary to look at and analyze failure cases. Derived from frequently misclassified samples in the testing sets, Figure \ref{fig:fc} shows some representative failure cases: 1) Poor illumination and light change environments with strong contrasts; 2) Special facial biometrics such as thicker eyebrows, heavier eye bags, and dark circles, which are rare in the training set; 3) Disturbances caused by decorations such as glasses and heavy make-up; 4) Common image variations like blurring and relatively tiny modification regions. Whether global or local, spatial or frequency, these strong perturbations bring challenges to feature extraction, resulting in difficulty to be distinguished.

\section{Conclusion}
In this paper, we introduce a novel method for face forgery detection, named MC-LCR, which models the correlation and interaction of local discrepancies. Given the challenges of small-scale tampered faces, high compression settings, and cross-dataset scenarios, the proposed framework is expected to effectively capture subtle forgery artifacts from both the spatial and frequency domains. To this end, SSRB is carefully devised to extract more discriminative local texture features, and PAPDA is developed to capture local inconsistencies in amplitude and phase spectrum. Meanwhile, the joint supervision of SC and CE loss is introduced to help disentangle the comprehensive representation from multi-modalities. The ablation studies and visualizations illustrated the effectiveness of each module. Extensive experiments demonstrate the robustness and superiority of our MC-LCR.

\section*{CRediT authorship contribution statement}
\textbf{Gaojian Wang:} Conceptualization, Methodology, Software, Investigation, Visualization, Formal analysis, Validation, Writing - Original Draft, Writing - Review \& Editing. \textbf{Qian Jiang:} Funding acquisition, Formal analysis,  Writing - Review \& Editing, Data Curation. \textbf{Xin Jin:} Writing - Review \& Editing, Validation, Visualization, Resources.
\textbf{Wei Li:} Writing - Review \& Editing, Conceptualization, Investigation, Validation.
\textbf{Xiaohui Cui:}  Writing - Review \& Editing, Project administration, Supervision, Resources.

\section*{Declaration of competing interes}
The authors declare that they have no known competing financial
interests or personal relationships that could have appeared
to influence the work reported in this paper.

\section*{ACKNOWLEDGMENT}
This study is supported by the National Natural Science Foundation of China (Nos. 61863036, 62002313, 62101481), Key Areas Research Program of Yunnan Province in China (No. 202001BB050076), and Key Laboratory in Software Engineering of Yunnan Province (No. 2020SE408).

\appendix


 \bibliographystyle{elsarticle-num} 
 \bibliography{cas-refs}





\end{document}